\documentclass[10pt,twocolumn,letterpaper]{article}

\usepackage[pagenumbers]{cvpr} 

\usepackage{times}
\usepackage{epsfig}
\usepackage{graphicx}
\usepackage{amsmath}
\usepackage{amssymb}
\usepackage{booktabs}

\usepackage{bm}
\usepackage{cases}
\usepackage[linesnumbered,ruled,vlined,lined,boxed,commentsnumbered]{algorithm2e}
\usepackage{indentfirst}
\usepackage{multirow}
\usepackage{boldline}
\usepackage{colortbl}

\usepackage{enumitem}
\usepackage{mathrsfs}
\usepackage{algorithmic}
\usepackage{bbding}
\usepackage{diagbox}

\usepackage[british,american]{babel}

\usepackage[dvipsnames]{xcolor}

\usepackage[pagebackref=true,breaklinks=true,colorlinks,bookmarks=false,citecolor=citecolor, linkcolor=linkcolor]{hyperref}
\definecolor{citecolor}{HTML}{0071BC}
\definecolor{linkcolor}{HTML}{ED1C24}

\makeatletter\renewcommand\paragraph{\@startsection{paragraph}{4}{\z@}
  {.5em \@plus1ex \@minus.2ex}{-.5em}{\normalfont\normalsize\bfseries}}\makeatother

\newcolumntype{x}[1]{>{\centering\arraybackslash}p{#1pt}}
\newlength\savewidth\newcommand\shline{\noalign{\global\savewidth\arrayrulewidth\global\arrayrulewidth 1pt}\hline\noalign{\global\arrayrulewidth\savewidth}}

\DeclareMathOperator*{\argmin}{arg\,min}

\usepackage[capitalize]{cleveref}
\crefname{section}{Sec.}{Secs.}
\Crefname{section}{Section}{Sections}
\Crefname{table}{Table}{Tables}
\crefname{table}{Tab.}{Tabs.}

\def\cvprPaperID{} 
\def\confName{CVPR}
\def\confYear{2022}

\definecolor{mborange}{rgb}{1.0,0.5,0.0}

\begin{document}

\title{
\vspace{-2mm}
Exploring Temporal Granularity in Self-Supervised \\Video Representation Learning
\vspace{-2mm}
}
\author{
Rui Qian$^{1,2}$\qquad
Yeqing Li$^{1}$\qquad
Liangzhe Yuan$^{1}$\qquad
Boqing Gong$^{1}$\qquad
Ting Liu$^{1}$ \\
Matthew Brown$^{1}$\qquad
Serge Belongie$^{3}$\qquad
Ming-Hsuan Yang$^{1}$\qquad
Hartwig Adam$^{1}$\qquad
Yin Cui$^{1}$ \\[2mm]
$^{1}$Google Research \qquad $^{2}$Cornell University \qquad $^{3}$University of Copenhagen\vspace{-2mm}}

\maketitle

\begin{abstract}
This work presents a self-supervised learning framework named \textbf{TeG} to explore \textbf{Te}mporal \textbf{G}ranularity in learning video representations.
In TeG, we sample a long clip from a video and a short clip that lies inside the long clip.
We then extract their dense temporal embeddings.
The training objective consists of two parts: a fine-grained temporal learning objective to maximize the similarity between corresponding temporal embeddings in the short clip and the long clip, and a persistent temporal learning objective to pull together global embeddings of the two clips.
Our study reveals the impact of temporal granularity with three major findings.
1) Different video tasks may require features of different temporal granularities. 
2) Intriguingly, some tasks that are widely considered to require temporal awareness can actually be well addressed by temporally persistent features.
3) The flexibility of TeG gives rise to state-of-the-art results on 8 video benchmarks, outperforming supervised pre-training in most cases.
\end{abstract}
\vspace{-2mm}

\section{Introduction}
\label{sec:intro}

Learning visual representations from abundantly available unlabeled videos is of crucial importance in computer vision, as the extra time dimension in videos enriches their content while significantly increasing the cost of manual annotation.
Thanks to the recent breakthroughs in image self-supervised learning~\cite{simclr,moco,byol,swav}, a series of more recent works extended similar ideas to videos~\cite{qian2021spatiotemporal,recasens2021broaden,feichtenhofer2021large}.

\begin{figure}[t]
  \centering
  \includegraphics[width=0.9\linewidth]{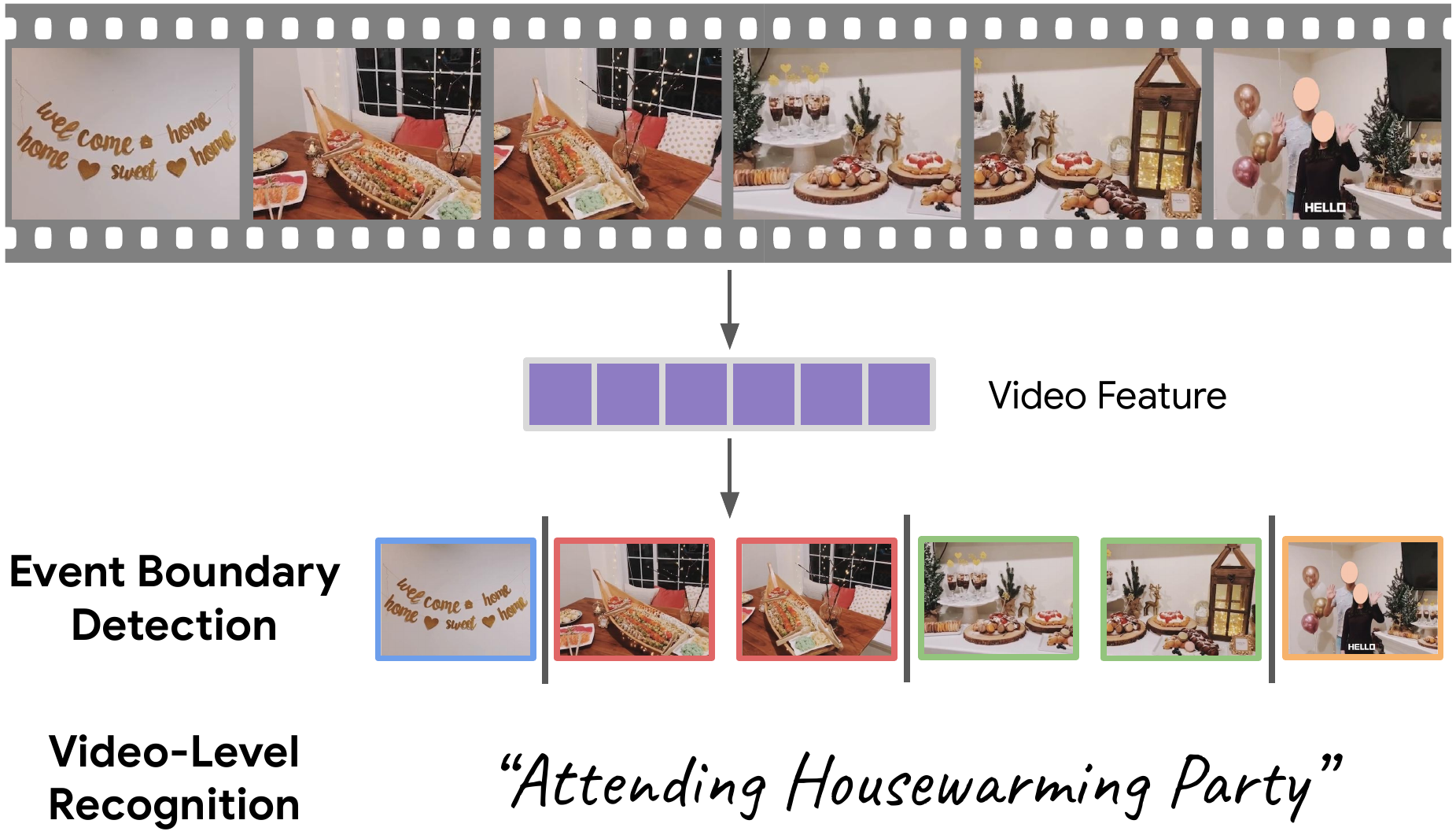}
\vspace{-2mm}
  \caption{Illustration of tasks requiring different temporal granularities on the same short video. Event boundary detection requires temporally fine-grained features within the video to achieve good performance, while in the case of video-level recognition, temporally persistent features are preferred.}
  \label{fig:teaser}
\vspace{-3mm}
\end{figure}

\begin{figure*}[t]
\centering
\includegraphics[width=0.9\textwidth]{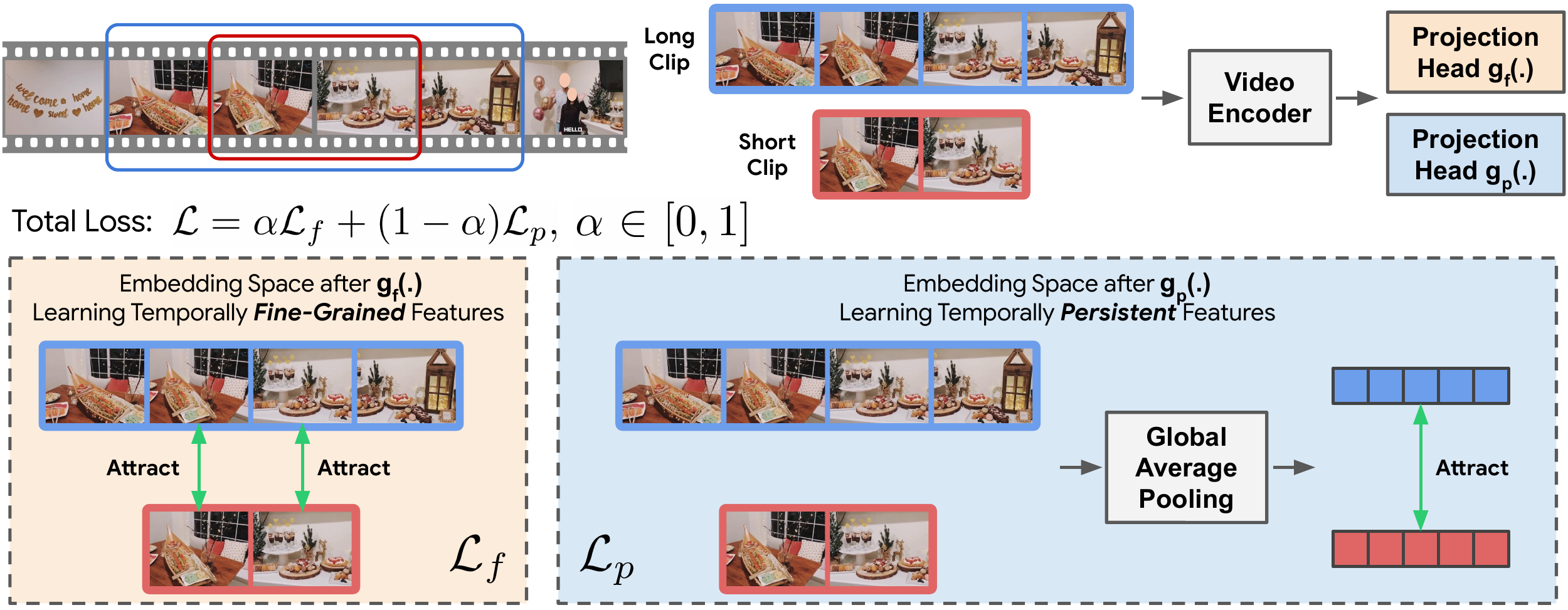}
\caption{Overview of our proposed TeG framework. We randomly sample a long clip from a given video and a short clip that lies inside the long clip. We then feed two clips into a video encoder without applying the final temporal average pooling. Two projection heads are applied to project features into separate embedding spaces, one for learning temporally fine-grained features and the other for learning temporally persistent features. For simplicity, spatial data augmentations and negative pairs from different videos are not illustrated.}
\label{fig:overview}
\vspace{-2mm}
\end{figure*}

The success of recent video self-supervised learning methods largely depends on a seemly counter-intuitive objective: enforcing temporal persistency across an entire video. 
More specifically, Qian~\etal~\cite{qian2021spatiotemporal} randomly sample two clips from a video as the positive pair to maximize their feature similarity; Recasens~\etal~\cite{recasens2021broaden} require the feature of a sampled clip to be close to the feature of the whole video in different modalities; Feichtenhofer~\etal~\cite{feichtenhofer2021large} pull together features from multiple clips within a video and observe that encouraging long temporal persistency can be effective even if the timespan is up to 60 seconds. 

Despite the strong performance on commonly used video benchmarks (\eg, action recognition~\cite{kay2017kinetics, ucf101, kuehne2011hmdb}), we find that features learned with such an objective do not perform well on more challenging video tasks proposed recently that require fine-grained temporal predictions~\cite{Sadhu_2021_CVPR,shou2021generic}.
This poses an interesting question: how can we develop a video self-supervised learning framework that accounts for both fine-grained and persistent temporal information?

We answer this question by rethinking video representation learning from the perspective of temporal granularity.
The concept of temporal granularity has been studied in speech recognition~\cite{fu2021scala} and time series analysis~\cite{costa2002multiscale, azami2019multivariate}, but is rather under-explored in the recent video representation learning research.
We find that different video tasks may require features of different temporal granularities.
In Figure~\ref{fig:teaser}, the event boundary detection calls for temporally fine-grained features so that the model is aware of the temporal content shifts.
In contrast, video-level recognition requires the model to robustly predict the target label based on some sampled clips from the video; therefore,  temporally persistent/coarse-grained features are more desirable.

The temporal dynamics of videos naturally provides a rich source of supervision for learning features with varying temporal granularities. 
As illustrated in Figure~\ref{fig:overview}, we propose \textbf{TeG}, a framework to explore \textbf{Te}mporal \textbf{G}ranularity via the combination of fine-grained and persistent temporal learning.
In TeG, we randomly sample a long clip from a video and a short clip that lies inside the time duration of the long clip. We then feed them into a video encoder without applying the final temporal average pooling. The resultant features are projected into two separate embedding spaces with different contrastive learning objectives.

In the fine-grained temporal learning space, for each clip, we split the projected features along the temporal dimension into a list of temporal embeddings, each represents the feature of a short time duration. Our sampling strategy guarantees that each temporal embedding in the short clip has a corresponding temporal embedding in the long clip at approximately the same start and end time. We apply a dense contrastive objective to maximize the similarity between corresponding temporal embeddings, which explicitly encourages temporal embeddings to be discriminative within a clip, making the learned features temporally fine-grained.

In the temporally persistent feature learning space, we directly apply a global average pooling to generate the global embedding for both the short clip and the long clip. 
The training objective here encourages global temporal persistency by pulling together two embeddings, similarly to what has been used in existing frameworks~\cite{qian2021spatiotemporal,recasens2021broaden,feichtenhofer2021large}. 
Our sampling strategy makes sure the global embedding from the long clip contains rich spatiotemporal context to avoid the model overly fitting to learning local features only. 

TeG optimizes both objectives and offers a flexible solution to learning features of different temporal granularities by adjusting the loss weight between the two objectives.

To understand the impact of the temporal granularity, we leverage two recently proposed datasets for understanding events in short videos to challenge self-supervised video representation learning: VidSitu event classification~\cite{Sadhu_2021_CVPR} and Kinetics-GEBD (generic event boundary detection)~\cite{shou2021generic}.
We conduct comprehensive experiments on the new benchmarks together with other commonly used video benchmarks.
We next summarize our main findings. 

First, \textbf{different video tasks may require features of different temporal granularities}.
For example, learning temporally fine-grained features improves the performance on VidSitu by 2.8\% and Kinetics-GEBD by 1.5\%, but hurts Kinetics linear evaluation  by 2.8\%. See Tables~\ref{tab:vidsitu_eval}, \ref{tab:gebd_eval} and \ref{tab:linear_eval}.

However, intriguingly, \textbf{temporally persistent features achieve strong performance on some video tasks that are widely considered requiring temporal awareness}.
Specifically, we achieve 61.4\% on Something-Something-v2 (SSv2) and 83.6\% on Diving48 with only the temporal persistency objective in TeG.
Our performance on Diving48 largely outperforms the supervised pre-training counterpart by 6.0\%.
Surprisingly, adding the fine-grained temporal learning objective yields a performance drop of 0.9\% on SSv2 and 2.1\% on Diving48.
See Tables~\ref{tab:ssv2-ft} and \ref{tab:diving48-ft}.

Finally, video representations learned with TeG advance the \textbf{state-of-the-art on a wide range of video tasks}.
In addition to the above mentioned results, TeG obtains a strong performance of 67.8\% on Kinetics linear evaluation. Fine-tuning the learned features on Kinetics yields 28.7 mAP on AVA-Kinetics, 94.1\% on UCF, and 71.9\% on HMDB. Our result on AVA-Kinetics significantly outperforms the supervised pre-training counterpart by 8.9 mAP. See Tables~\ref{tab:linear_eval}, \ref{tab:avak-ft} and \ref{tab:ucf}.
Other than classic video benchmarks, we achieve 31.1\% accuracy on VidSitu, which is on par with the supervised pre-training. On Kinetics-GEBD, TeG obtains 71.4\% F1 score, outperforming a strong supervised pre-training method by a large margin of 8.9\%.
See Tables~\ref{tab:vidsitu_eval} and \ref{tab:gebd_eval}.

\section{Related Work}
\label{sec:related_work}

\paragraph{Unsupervised video representation learning.}
The temporal dimension of video has been densely explored. 
In an early work, Srivastava~\etal~\cite{srivastava2015unsupervised} propose to predict the future based on frame features. More recent work learn from raw videos by predicting motion and appearance statistics~\cite{wang2019self}, speed~\cite{benaim2020speednet, wang2020self} and encodings~\cite{lotter2016deep,han2019video,han2020memory}. 
Aside from future prediction, it is common to learn from pretext tasks like sorting frames or video clips~\cite{lee2017unsupervised,xu2019self,kim2019self,fernando2017self} and rotation~\cite{jing2018self}. Recently, constrastive learning based methods~\cite{singh2021semi, qian2021spatiotemporal, feichtenhofer2021large, recasens2021broaden} significantly reduce the gap with supervised learning by pulling together features of clips from the same video. 
Furthermore, videos containing multimodal signals make it possible to learn from cross-modality self-supervision, \eg geometric cues~\cite{gan2018geometry}, speech or language~\cite{sun2019videobert, sun2019learning, miech2020end}, audio~\cite{korbar2018cooperative, alwassel2019self, patrick2020multi, asano2020labelling}, optical flow~\cite{han2020coclr}, or combinations of multiple modalities~\cite{alayrac2020self, recasens2021broaden, akbari2021vatt} and tasks~\cite{Piergiovanni_2020_CVPR}.
Different from existing work, we introduce temporally fine-grained features into the video contrastive learning framework and study its impact on various downstream tasks.

\paragraph{Fine-grained temporal video understanding.}
We first discuss two representative tasks: temporal localization and segmentation.
Commonly used temporal localization benchmarks (\eg, ActivityNet~\cite{activitynet}, THUMOS~\cite{thumos}, HACS~\cite{zhao2019hacs}) are constructed based on specified action classes. 
As a result, most temporal localization methods~\cite{shou2016temporal, shou2017cdc, zhao2017temporal, lin2018bsn, lin2019bmn, long2019gaussian} contain a temporal proposal module to simply treat video segments that do not belong to pre-defined classes as the background. 
Temporal segmentation methods~\cite{lea2016segmental, richard2017weakly, ding2018weakly} typically divide a video into segments of actions, or sub-actions~\cite{shao2020intra, shao2020finegym}.
But still, those methods can only predict boundaries of pre-defined classes, not generic boundaries.
On the other hand, studies in cognitive science~\cite{tversky2013event} show that humans naturally segment videos into meaningful temporal chunks of events, without pre-defined categories.
Inspired by this, we choose the recently proposed Kinetics-GEBD~\cite{shou2021generic} dataset to verify whether TeG is able to learn temporally fine-grained features that can be used for generic event boundary detection.
We also benchmark our method on AVA-Kinetics~\cite{li2020ava}, a classic spatiotemporal action localization dataset.
In addition to videos containing human actions, movies could also provide rich content for fine-grained temporal video understanding. 
However, temporal movie understanding methods~\cite{huang2020movienet, chen2021shot, pardo2021learning} typically focus on shots, which are defined by sharp transitions due to video editing and can be accurately localized using low-level visual cues~\cite{sidiropoulos2011temporal}.
To better benchmark our method in complex movie scenes, we adopt the recently proposed VidSitu~\cite{Sadhu_2021_CVPR} dataset.
In VidSitu, each short video is temporally annotated with 5 events. Transitions between events are usually natural and continuous, and thus cannot be detected by low-level visual cues.

\section{Method}
\label{sec:method}

Our proposed framework is illustrated in Figure~\ref{fig:overview}. 
We next introduce each component in detail. 

\paragraph{Temporal sampling.} 
Given a video of $N$ frames, $V = \{v_1, v_2, \cdots, v_N\}$, previous works~\cite{qian2021spatiotemporal, feichtenhofer2021large} typically sample two short clips with the same length independently from $V$ aiming at learning temporally persistent features. 
However, this common strategy is not suitable for learning temporally fine-grained features since it enforces non-overlapped clips to have similar features. 
Sampling two clips that have some overlaps would partially avoid this issue, but it sacrifices diversity in spatiotemporal context, resulting in inferior representations.
Hence, we propose a long-short sampling strategy, where we first sample a long clip $l$ randomly from the whole video, and then we sample a short clip $s$ from the whole video inside the time duration of the long clip.
The long clip provides rich spatiotemporal context, and the  short clip in it ensures the existence of corresponding embeddings between the two clips. The ablation on sampling strategy is in Table~\ref{tab:ablation_sample}.

We note that some very recent works~\cite{recasens2021broaden, wang2021long} also propose an asymmetric sampling strategy. 
Our method differs from them in the following key aspects: 
1) Motivation: we propose this design to find corresponding embeddings and preserve rich spatiotemporal context at the same time, while Recasens~\etal~\cite{recasens2021broaden} aim at accommodating different modalities, and Wang~\etal~\cite{wang2021long} focus on pre-training for video transformers. 
2) Objective: we use this design to learn temporally fine-grained features. In contrast, their works~\cite{recasens2021broaden, wang2021long} still emphasize the invariance across the whole video. 
3) Implementation: based on our motivation and objective, we require the short clip to fall inside the long clip, while they~\cite{recasens2021broaden, wang2021long} perform independent sampling which is similar to other video self-supervised learning methods using symmetric sampling~\cite{qian2021spatiotemporal, feichtenhofer2021large}. 
Finally, we conduct dense contrastive learning between corresponding embeddings, which is fundamentally different from their work~\cite{recasens2021broaden, wang2021long}.

\paragraph{Spatial data augmentation.} 
After obtaining the short clip $s$ and long clip $l$, we adopt the common practice in recent video contrastive learning~\cite{qian2021spatiotemporal, alayrac2020self, alwassel2019self} and apply a series of spatial data augmentations including random resizing and cropping, color jittering, and Gaussian blurring. The parameters of the spatial augmentations follow Qian~\etal~\cite{qian2021spatiotemporal}.

\paragraph{Video encoder.} 
We adopt the 3D-ResNet-50 (R3D-50) backbone architecture used in~\cite{qian2021spatiotemporal}. We remove the final \emph{temporal} average pooling and only keep the global \emph{spatial} average pooling since we are primarily interested in exploring \emph{temporal} granularities. We notate the modified encoder as $f(\cdot)$. 
We apply two projection heads: $g_p(\cdot)$ for persistent temporal learning and $g_f(\cdot)$ for fine-grained temporal learning. They project representations into separate embedding spaces with different contrastive objectives. 
In the persistent learning space, we obtain embedding $z^s_p$ from the short clip input $s$ and $z^l_p$ from the long clip input $l$ by $\{z^s_p, z^l_p\} = \{g_p (f(s)), g_p (f(l))\}$; in the fine-grained learning space, we have $\{z^s_f, z^l_f\} = \{g_f (f(s)), g_f (f(l))\}$.

Our approach maintains a simple form of video contrastive learning where we do not use separate encoders for different clips~\cite{recasens2021broaden}, nor do we use a momentum encoder~\cite{feichtenhofer2021large}, predictor head~\cite{feichtenhofer2021large, wang2021long} and symmetric losses~\cite{recasens2021broaden}. 
Extensive experiments in Section~\ref{sec:experiments} demonstrate the effectiveness of this simple design. 

\paragraph{Temporal aggregation.} 
For temporally persistent learning, as a common practice~\cite{qian2021spatiotemporal, feichtenhofer2021large}, we directly apply a global average pooling along the temporal dimension to get a single vector representing the whole clip, resulting in $z^s_p, z^l_p \in \mathbb{R}^{1 \times c}$, where $c$ is the number of output channels from the projection head. 
For temporally fine-grained learning, we design a configurable local aggregation strategy to optionally aggregate consecutive local temporal embeddings to reduce training complexity. 
We denote the number of frames in short clip $s$ and long clip $l$ as $T_s$ and $T_l$, respectively. 
Our aggregation strategy performs average pooling on every consecutive $\frac{T_s}{n}$ frames in the short clip and every consecutive $\frac{T_l}{m}$ frames in the long clip, resulting in aggregated outputs of $z^s_f \in \mathbb{R}^{n \times c}$ and $z^l_f \in \mathbb{R}^{m \times c}$. 
When $n = 1$ and $m = 1$, it reduces to temporal persistent learning. 
When $n = T_S$ and $m = T_L$, it conducts dense temporal contrastive learning on frame-level embeddings. 
Figure~\ref{fig:ablation_nm} ablates the effect of different choices of $n$ and $m$. 
We further use $z^s_f [i]$ to index the $i$-th dimension of $z^s_f$ and $z^l_f [j]$ to index the $j$-th dimension of $z^l_f$, where $1 \leq i \leq n$ and $1 \leq j \leq m$.

\paragraph{Fine-grained temporal learning.} 
We aim to obtain temporally fine-grained features by maximizing the feature similarity between corresponding embeddings of the short and the long clip. 
The corresponding embeddings should be close in time and we rely on the frame index to find them. Since we conduct temporal aggregation on a few consecutive frames, we define the index of a certain embedding $z^s_f [i]$ after aggregation as the average frame index of all aggregated frames, notated as $I(z^s_f [i])$. We find $z^s_f [i]$'s corresponding embedding $z^l_f [j]$ in the long clip by:
\begin{equation}
    j = \argmin_j | I(z^s_f [i]) - I(z^l_f [j])|.
\end{equation}
$(z^s_f [i], z^l_f [j])$ has the closest temporal distance and it is considered as the positive pair. The fine-grained temporal learning loss can be written as:
\begin{equation}
\small
    \mathcal{L}_f = - \frac{1}{n} \sum\limits_{i=1}^{n} \log\frac{
    \exp(z^s_f [i] \cdot z^l_f [j] / \tau)
    }{
    \exp(z^s_f [i] \cdot z^l_f [j] / \tau)
    + \sum\limits_{k_f^-} \exp (z^s_f [i] \cdot k_f^- / \tau))
    },
    \label{eq:temp}
\end{equation}
where $k_f^-$ represents all dense embeddings of long clips from other videos after temporal aggregation in the fine-grained temporal learning space and $\tau$ is the temperature.

\paragraph{Persistent temporal learning.} 
Recall that we have embeddings $z^s_p, z^l_p \in \mathbb{R}^{1 \times c}$ in the temporally persistent learning space. $(z^s_p, z^l_p)$ is considered as the positive pair and $(z^s_p, k_p^-)$ are considered as negative pairs, where $k_p^-$ represents all global embeddings from long clips of other videos in the persistent temporal learning space. The persistent temporal learning loss can be written as:
\begin{equation}
    \mathcal{L}_p=-\log\frac{\exp(z^s_p \cdot z^l_p / \tau)}{\exp(z^s_p \cdot z^l_p / \tau) + \sum\limits_{k_p^-} \exp(z^s_p \cdot k_p^- / \tau ) }.
    \label{eq:clip}
\end{equation}
$\mathcal{L}_p$ adopts the same temperature $\tau$ with $\mathcal{L}_f$ for simplicity.

\paragraph{Total loss.}
The total loss is a weighted sum of the fine-grained and persistent temporal learning loss:
\begin{equation}
    \mathcal{L} = \alpha \mathcal{L}_f + (1 - \alpha) \mathcal{L}_p, 
    \label{eq:final}
\end{equation}
where the weight $\alpha \in [0,1]$ is used to control the temporal granularity of the learned features. When $\alpha$ is close to 0, we intend to learn temporally persistent features with only $\mathcal{L}_p$ in the loss. With the increasing of $\alpha$, more temporally fine-grained features will be obtained. An ablation regarding the effect of $\alpha$ on two datasets is presented in Figure~\ref{fig:ablation_alpha}.

\section{Evaluation}
\label{sec:evaluation}

To evaluate our proposed framework, we leverage two recently proposed datasets: VidSitu~\cite{Sadhu_2021_CVPR} for event classification and Kinetics-GEBD~\cite{shou2021generic} for generic event boundary detection. 
Additionally, we also evaluate on 6 commonly used datasets, including Kinetics via linear probing and various downstream tasks via fine-tuning. See Section~\ref{sec:experiments} for more details. 
We next describe how we evaluate our method on these 2 new datasets.

\paragraph{Event classification.} 
VidSitu~\cite{Sadhu_2021_CVPR} is a large-scale movie dataset focusing on understanding the relationship of events in short videos. 
Each video in VidSitu is 10-second long and divided into 5 consecutive and non-overlapping events. Each event is annotated with a verb to describe the most salient action taking place inside it. The temporal duration of the events is determined by human perception to avoid including multiple events. 
The baseline provided by the original authors is to first cut the video into 5 events according to the annotated boundaries and then perform classification for each event. 
In our case, we directly apply our method on raw videos in VidSitu without using any labels in pre-training. 
An interesting property of this dataset is that the transition between events is usually natural and continuous, and cannot be easily detected by an off-the-shelf shot detector~\cite{sidiropoulos2011temporal}.
Thus we consider VidSitu a good benchmark to evaluate whether our method can learn more fine-grained temporal features than current state-of-the-art video-level persistent learning methods~\cite{qian2021spatiotemporal}. 
We adopt linear probing in the evaluation, where we use their event labels to train a linear classifier on top the frozen backbone to quantify the performance of the learned representations.  

\paragraph{Generic event boundary detection.}
Kinetics-GEBD~\cite{shou2021generic} annotates Kinetics-400~\cite{kay2017kinetics} videos with fine-grained event boundaries based on human perception. 
The boundaries are in the format of timestamps and a detection is considered as correct when its temporal distance with a ground truth is smaller than 5\% of the total video length. 
We use a 1D sliding window detection method to detect event boundaries, following the spirit of classic object detection methods like HOG~\cite{dalal2005histograms} and DPM~\cite{felzenszwalb2009object}.
We first pre-train our backbone without using any annotations. 
We then add a binary linear classifier on top of the pre-trained backbone to predict whether a clip contains a boundary or not.
Similar to object detection~\cite{girshick2014rich,ren2015faster}, we fine-tune the model end-to-end to verify the performance of our learned features.

\section{Experiments}
\label{sec:experiments}

We first introduce general implementation details, followed by task-specific settings and experimental results.

\subsection{Implementation Details}
\label{sec:experiments_implementation}
We use SGD with the momentum of 0.9 as our optimizer. During the self-supervised pre-training, we follow~\cite{qian2021spatiotemporal} to train models with 1024 batch size and 0.32 learning rate, using linear warm up in the first 5 epochs~\cite{goyal2017accurate} followed by half-period cosine decay~\cite{he2019bag}. The temperature $\tau$ in loss function is set to 0.156 for pre-training on Kinetics, and 0.1 for all other 
datasets.
We adopt two representative settings in our method:
1) $\alpha = 0.0$ for persistent temporal learning only and we call this method \textbf{TeG-PS}, where PS represents ``persistent''. 
2) $\alpha = 0.9$, in which the fine-grained temporal learning loss is the dominant loss and we denote this method as \textbf{TeG-FG}, where FG represents ``fine-grained''. 
Our code and models will be made available to the public.

\subsection{Event Classification} 
We conduct experiments on VidSitu~\cite{Sadhu_2021_CVPR}, which contains 23.6k training and 1.3k validation videos from 1560 verb classes. 
Each video is exactly 10-second long and is divided into 5 event clips. Each event clip in the validation set receives 10 annotations to respect the flexibility of describing an event with multiple similar verbs. During pre-training, we sample a 32-frame long clip with a stride of 4 and a 16-frame short clip with a stride of 2. 
Temporal aggregation parameters are set as $m=4$ and $n=1$ (ablation study in Figure~\ref{fig:ablation_nm}). 
We pre-train our model from scratch for 200 epochs on unlabeled raw videos. 
During the linear evaluation, we train a linear classifier with an initial learning rate of 4.0 for 100 epochs. To deal with multiple human annotations in the validation set, we follow the practice from the original authors~\cite{Sadhu_2021_CVPR} by only considering the set of classes that appear at least twice within the 10 annotations. A prediction would be considered as correct if it matches any class in the set.
Additional details can be found in Appendix~\ref{app_vidsitu}.

We show TeG's performance on VidSitu in Table~\ref{tab:vidsitu_eval}. 
The supervised methods directly train models from scratch on the training set, using labels for each event clip cut from raw videos. 
The unsupervised methods perform pre-training on raw videos from scratch without using any labels and then conduct linear evaluation. 
We adopt CVRL~\cite{qian2021spatiotemporal} as an important baseline since it is a representative method that enforces temporal persistency across the whole video. 
TeG-PS achieves similar performance with CVRL while TeG-FG equipped with temporally fine-grained pre-training improves the performance by 2.8\%. 
Furthermore, the performance TeG-FG is on par with supervised methods using I3D as the backbone. This result provides a solid evidence that temporal persistent learning is not the optimal solution on this event classification benchmark.

\begin{table}[h]
\small
\centering
\begin{tabular}{clrr}
\multicolumn{2}{c}{method}& \multicolumn{1}{c}{backbone} & \multicolumn{1}{r}{acc.} \\
\shline
\multirow{4}{*}{Supervised}& \multirow{4}{*}{\begin{tabular}[c]{@{}c@{}}Train from\\ scratch~\cite{Sadhu_2021_CVPR}\end{tabular}} & \multicolumn{1}{r}{I3D} & \multicolumn{1}{r}{31.2} \\
& & \multicolumn{1}{r}{I3D + NL} & \multicolumn{1}{r}{30.2} \\
& & R3D-50 + NL & 33.1\\
& & SlowFast + NL & 32.6 \\
\hline
\multicolumn{1}{c}{\multirow{3}{*}{Unsupervised}} & \multicolumn{1}{l}{CVRL~\cite{qian2021spatiotemporal}} & R3D-50 & 28.3 \\
\multicolumn{1}{l}{} & \multicolumn{1}{l}{TeG-PS} & R3D-50 & 28.3 \\
\multicolumn{1}{l}{} & \multicolumn{1}{l}{TeG-FG} & R3D-50 & \textbf{31.1}\\
\end{tabular}
\caption{\textbf{Event classification on Vidsitu.} TeG-FG with fine-grained temporal learning outperforms persistent temporal learning and is on par with supervised methods.}
\vspace{-4mm}
\label{tab:vidsitu_eval}
\end{table}

\subsection{Generic Event Boundary Detection} 
We perform experiments on Kinetics-GEBD~\cite{shou2021generic}, which contains 20k out of 240k Kinetics-400~\cite{kay2017kinetics} training videos and all 20k validation videos, each annotated with 5 sets of event boundaries. 
We sample a 16-frame long clip and a 8-frame short clip, both with a stride of 2. Temporal aggregation parameters are set as $m=2$ and $n=1$. 
We pre-train our model from scratch for 200 epochs and then fine-tune the model with the annotated boundaries for 30 epochs. The temporal sliding window is set with a duration of 1.28s and a stride of 0.12s to go through all videos to generate all clips. The clips are considered as positive when the timestamp of the clip's center is within 0.15s of the annotated boundaries, following the original authors~\cite{shou2021generic}. 
This results in 0.2M positive and 1.2M negative clips in total. Balanced sampling is applied for each training batch to avoid the model overly focusing on negative examples. 
For evaluation, we compare our prediction with all annotations in the same video and select the annotation with the highest F1 score as the ground-truth, which is the standard practice in the official evaluation code provided the authors~\cite{shou2021generic}. 
Additional details can be found in Appendix~\ref{app_gebd}.

TeG's performance on Kinetics-GEBD is presented in Table~\ref{tab:gebd_eval}, where we report results using their strictest temporal threshold of 0.05 to emphasize on the importance of precise boundary detection. 
We first briefly introduce a few representative methods for this benchmark. SceneDet~\cite{pyscenedetect} is a widely-used library for detecting shot changes. BMN~\cite{lin2019bmn} is a state-of-the-art method for action proposal generation and here the start and end of each proposal are considered as event boundaries. 
The dataset creators~\cite{shou2021generic} also develop an improved version of BMN called BMN-SE. TCN~\cite{lea2016segmental} is a classic action boundary detection method. 
PC~\cite{shou2021generic} is the state-of-the-art method on this benchmark provided by performing pairwise classification around event boundaries. We group these methods by the external data they pre-train on and whether they fine-tune or keep the backbone frozen.
From results in Table~\ref{tab:gebd_eval}, fine-tuning methods clearly achieve much better performances. 
Compared to PC which relies on ImageNet supervised pre-training, CVRL~\cite{qian2021spatiotemporal} and TeG can directly pre-train on the training videos without using labels and external data for supervision. 
We draw a similar observation with event classification that TeG-FG with temporally fine-grained learning outperforms methods enforcing temporal persistency like CVRL and TeG-PS.

\begin{table}[h]
\small
\centering
\begin{tabular}{lrcr}
method & external data & finetuning & F1\\
\shline
BMN~\cite{lin2019bmn} & IN + THUMOS & \tiny{\XSolidBrush} & 18.6 \\
SceneDet~\cite{pyscenedetect} & - & \tiny{\XSolidBrush} & 27.5 \\
PA~\cite{shou2021generic} & IN & \tiny{\XSolidBrush} & 39.6 \\ 
BMN-SE~\cite{lin2019bmn} & IN + THUMOS & \tiny{\XSolidBrush} & 49.1 \\
TCN~\cite{lea2016segmental} & IN & \tiny{\XSolidBrush} & 58.8 \\
\hline
PC~\cite{shou2021generic} & IN & \checkmark & 62.5 \\
CVRL~\cite{qian2021spatiotemporal} & - & \checkmark & 69.1 \\
TeG-PS & - & \checkmark & 69.9 \\
TeG-FG & - & \checkmark & \textbf{71.4} \\
\end{tabular}
\caption{\textbf{Event boundary detection on Kinetics-GEBD.} IN represents ImageNet supervised pre-training and THUMOS means additional supervised training on THUMOS~\cite{thumos}. TeG-FG with fine-grained temporal learning shows better performance then persistent temporal learning methods TeG-PS and CVRL. TeG-FG also surpass PC with ImageNet supervised pre-training by 8.9\%.}
\vspace{-2mm}
\label{tab:gebd_eval}
\end{table}

\subsection{Kinetics Linear Evaluation}
We pre-train our model from scratch for 800 epochs on Kinetics-400~\cite{kay2017kinetics}.
We use the same parameters with event classification: the long clip of 32-frame (4-stride), the short clip of 16-frame (2-stride) and the temporal aggregation of $m=4, n=1$.
We perform linear evaluation which we consider as the most straightforward way to quantify the learned feature quality, following the same setting in~\cite{qian2021spatiotemporal}. 
As shown in Table~\ref{tab:linear_eval}, TeG-FG obtains 65.0\% top-1 accuracy which trails behind some state-of-the-art methods such as CVRL~\cite{qian2021spatiotemporal} and $\rho$MoCo~\cite{feichtenhofer2021large}. 
By contrast, TeG-PS achieves 67.8\%, which is state-of-the-art on Kinetics linear evaluation without using multi-clip sampling~\cite{feichtenhofer2021large} in pre-training. This verifies that temporal persistency in the key to obtain strong performance on Kinetics. 
We believe the performance of TeG-PS could be additionally boosted by multi-clip sampling as it further enhances temporal persistency.

\begin{table}[t]
\small
\centering
\begin{tabular}{lrrr}
method & backbone & pre-train data & acc. \\
\shline
VTHCL~\cite{yang2020video} & R3D-50 & K400 & 37.8 \\
SimCLR~\cite{qian2021spatiotemporal} & R3D-50 & K400 & 46.8 \\
VINCE~\cite{gordon2020watching} & R-50 & K400 & 49.1 \\
ImageNet~\cite{qian2021spatiotemporal} & R3D-50 & IN & 53.5\\
SeCo~\cite{yao2020seco} & R-50 & IN$^\dagger$ + K400 & 61.9 \\
$\rho$SwAV ($\rho$=2)~\cite{feichtenhofer2021large} &  R3D-50 & K400 & 63.2\\
CVRL~\cite{qian2021spatiotemporal} & R3D-50 & K400 & 66.1\\
$\rho$BYOL ($\rho$=2)~\cite{feichtenhofer2021large} &  R3D-50 & K400 & 66.2\\
MCL~\cite{li2021motion} &  R(2+1)D-50 & IN$^\dagger$ + K400 & 66.6 \\
$\rho$MoCo ($\rho$=2)~\cite{feichtenhofer2021large} &  R3D-50 & K400 & 67.4\\
\hline
TeG-FG & R3D-50 & K400 & 65.0\\
TeG-PS & R3D-50 & K400 & \textbf{67.8}\\
\end{tabular}
\caption{\textbf{Linear evaluation on Kinetics-400 action recognition.} For fair comparison, we quote results in~\cite{feichtenhofer2021large} under the same setting with $\rho$=2 (sampling two clips from a video) and 800 epochs of pre-training. IN$^\dagger$ denotes a MoCo-v2~\cite{chen2020improved} checkpoint pre-trained on ImageNet is used as the initialization of the backbone.}
\vspace{-2mm}
\label{tab:linear_eval}
\end{table}

\subsection{Downstream Action Recognition} 
For the downstream action recognition task, we fine-tune the same pre-trained checkpoint from Kinetics linear evaluation on four benchmarks: 
Something-Something-v2 (SSv2)~\cite{goyal2017something}, Diving48~\cite{li2018resound}, UCF101~\cite{ucf101} and HMDB51~\cite{kuehne2011hmdb}. 
We attach a linear layer with a dropout rate of 0.5 after the backbone and train the model end-to-end for 50 epochs on all these benchmarks, with different learning rates of 3.0 for SSv2, 2.0 for Diving48, 0.64 for UCF and 0.32 for HMDB. 

Something-Something-v2 and Diving48 are two representative datasets that are widely considered requiring temporal awareness and we first study the transfer performance on them. 
We find the two benchmarks in fact can be well addressed by solely learning temporally persistent features and bringing in temporally fine-grained features is detrimental. 
Concretely, as shown in Table~\ref{tab:ssv2-ft}, TeG-PS achieves 61.4\% accuracy, surpassing TeG-FG as well as all other state-of-the-art unsupervised pre-training methods on SSv2. More surprisingly, TeG-PS, using a small R3D-50 backbone, is 1.9\% better than TimeSformer and on par with SlowFast with supervised pre-training.

\begin{table}[htbp]
\small
\centering
\begin{tabular}{clrr}
\multicolumn{2}{c}{method} & pre-train data & acc. \\
\shline
\multirow{4}{*}{\begin{tabular}[c]{@{}c@{}}Sup. \\ pre-train\end{tabular}} 
& TimeSformer~\cite{bertasius2021space} & ImageNet & 59.5\\
& SlowFast~\cite{slowfast} & K400 & 61.7\\
& TimeSformer-L ~\cite{bertasius2021space} & ImageNet & 62.0\\
& TSM~\cite{tsm} & K400 & 63.3\\
\hline
\multirow{5}{*}{\begin{tabular}[c]{@{}c@{}}Unsup. \\ pre-train\end{tabular}} 
& $\rho$MoCo ($\rho$=2)~\cite{feichtenhofer2021large} & K400 & 54.4\\
& $\rho$BYOL ($\rho$=2)~\cite{feichtenhofer2021large} & K400 & 55.8\\
& CVRL~\cite{qian2021spatiotemporal} & K400 & 59.6 \\
& TeG-FG & K400 & 60.5\\
& TeG-PS & K400 & \textbf{61.4}\\
\end{tabular}
\caption{\textbf{Action recognition on Something-Something-v2.} Despite the dataset is considered requiring more temporal awareness, TeG-PS outperforms TeG-FG, and is on par with strong supervised pre-training methods.}
\vspace{-4mm}
\label{tab:ssv2-ft}
\end{table}

We observe similar trends from the results on Diving48 in Table~\ref{tab:diving48-ft}, where the advantage of TeG-PS is further enhanced. 
TeG-PS achieves 83.6\%, which is more than 2\% better than TeG-FG and CVRL. 
Note that we adopt the updated annotations (Oct. 2020)\footnote{\href{http://www.svcl.ucsd.edu/projects/resound/dataset.html}{http://www.svcl.ucsd.edu/projects/resound/dataset.html}}
provided by the original author, where the mislabelled data are corrected. For fair comparison, all results we report here are conducted on the updated annotations.
To our surprise, simply using unsupervised pre-training and a R3D-50 backbone, TeG-PS is able to obtain significantly better performance than state-of-the-art supervised pre-training methods using much stronger backbones, \eg SlowFast~\cite{slowfast}, TimeSformer and TimeSformer-L~\cite{bertasius2021space}.

\begin{table}[h]
\small
\centering
\begin{tabular}{clrr}
\multicolumn{2}{c}{method} & pre-train data & acc. \\
\shline
\multirow{6}{*}{\begin{tabular}[c]{@{}c@{}}Sup. \\ pre-train\end{tabular}} & I3D~\cite{i3d} & K400 & 48.3 \\
& TSM~\cite{tsm} & ImageNet & 51.1\\
& TSN~\cite{wang2016temporal} & ImageNet & 52.5\\
& TimeSformer~\cite{bertasius2021space} & ImageNet & 74.9\\
& SlowFast~\cite{slowfast} & K400 & 77.6\\
& TimeSformer-L~\cite{bertasius2021space} & ImageNet & 81.0\\
\hline
\multirow{3}{*}{\begin{tabular}[c]{@{}c@{}}Unsup. \\ pre-train\end{tabular}} & CVRL~\cite{qian2021spatiotemporal} & K400 & 80.9 \\
& TeG-FG & K400 & 81.5\\
& TeG-PS & K400 & \textbf{83.6}\\
\end{tabular}
\caption{\textbf{Action recognition on Diving48.} TeG-PS sets a new record of 83.6\% accuracy, outperforming all unsupervised and supervised pre-training methods.}
\vspace{-2mm}
\label{tab:diving48-ft}
\end{table}

We then report TeG's performance on UCF101 and HMDB51 in Table~\ref{tab:ucf}, which are classic benchmarks for evaluating self-supervised video representation learning. 
On UCF, TeG-PS achieves state-of-the-art performance of 94.1\% with fine-tuning and 91.1\% with linear evaluation. On HMDB, TeG-PS achives 71.9\% with fine-tuning which surpasses CVRL by 4\% and 64.2\% with linear evaluation which is 5.9\% better than CVRL.
 
\begin{table}[t]
\small
\centering
\begin{tabular}{lrrr}
method & pre-train data & UCF & HMDB \\
\shline
\multicolumn{4}{c}{\multirow{1}{*}{\small{\textbf{Fine-Tuning}}}} \\
MotionPred~\cite{wang2019self} & K400 & 61.2 & 33.4 \\
3D-RotNet~\cite{jing2018self} & K400 & 64.5 & 34.3 \\
ST-Puzzle~\cite{kim2019self} & K400 & 65.8 & 33.7 \\
ClipOrder~\cite{xu2019self} & K400 & 72.4 & 30.9 \\
DPC~\cite{han2019video} & K400 & 75.7 & 35.7 \\
PacePred~\cite{wang2020self} & K400 & 77.1 & 36.6 \\
MemDPC~\cite{han2020memory} & K400 & 78.1 & 41.2 \\
SpeedNet~\cite{benaim2020speednet} & K400 & 81.1 & 48.8\\
CoCLR~\cite{han2020coclr} & K400 & 87.9 & 54.6 \\
DynamoNet~\cite{diba2019dynamonet} & YT8M & 88.1 & 59.9 \\
SeCo~\cite{yao2020seco} & IN$^\dagger$ + K400 & 88.3 & 55.6\\
CVRL~\cite{qian2021spatiotemporal} & K400 & 92.9 & 67.9\\
$\rho$MoCo ($\rho$=2)~\cite{feichtenhofer2021large} & K400 & 93.2 & - \\
MCL~\cite{li2021motion} & IN$^\dagger$ + K400 & 93.4 & 69.1 \\
\hline
TeG-FG & K400 & 93.6 & 70.7 \\
TeG-PS & K400 & \textbf{94.1} & \textbf{71.9}\\
\multicolumn{4}{c}{\multirow{1}{*}{\small{\textbf{Linear Evaluation}}}} \\
MemDPC~\cite{han2020memory} & K400 & 54.1 & 30.5 \\
CoCLR~\cite{han2020coclr} & K400 & 77.8 & 52.4 \\
CVRL~\cite{qian2021spatiotemporal} & K400 & 89.8 & 58.3 \\
\hline
TeG-FG & K400 & 88.9 & 60.7 \\
TeG-PS & K400 & \textbf{91.1} & \textbf{64.2}\\
\end{tabular}
\caption{\textbf{Action recognition on UCF101 and HMDB51.} 
TeG-PS shows superior performance over other methods. IN$^\dagger$ denotes a MoCo-v2~\cite{chen2020improved} checkpoint pre-trained on ImageNet is used as the initialization of the backbone.} 
\vspace{-2mm}
\label{tab:ucf}
\end{table}

\subsection{Downstream Action Localization}
AVA-Kinetics~\cite{li2020ava} provides an important spatiotemporal action localization benchmark for evaluating the learned video features. 
We use our pre-trained backbone to extract features from the given clip. Following the practice in recent work~\cite{li2020ava, girdhar2019video}, we use an action transformer head to capture the relationship between each person and the whole scene, and conduct training using ground truth boxes and evaluate on person detections provided by an off-the-shelf detector~\cite{zhou2019objects}. We fine-tune the whole network for 36 epochs with an initial learning rate of 0.01 and a batch size of 256.

The results are shown in Table~\ref{tab:avak-ft}, where TeG-PS achieves 28.7 mAP, outperforming supervised pre-training on Kinetics using the same R3D-50 backbone by a large margin of 8.9 mAP. TeG-PS also shows superior performance when compared with other state-of-the-art unsupervised pre-training methods like CVRL and VFS~\cite{xu2021rethinking}. 
TeG-FG is 1.0 mAP lower than TeG-PS, indicating that learning temporally fine-grained feature is not helpful on this task.

\begin{table}[h]
\small
\centering
\begin{tabular}{clrr}
\multicolumn{2}{c}{method} & pre-train data & mAP \\
\shline
\multirow{2}{*}{\begin{tabular}[c]{@{}c@{}}Sup. \\ pre-train\end{tabular}}
& R3D-50 & K400 & 19.8 \\
& I3D~\cite{i3d} & ImageNet & 22.9 \\
\hline
\multirow{4}{*}{\begin{tabular}[c]{@{}c@{}}Unsup. \\ pre-train\end{tabular}}
& CVRL~\cite{qian2021spatiotemporal} & K400 & 24.1 \\
& VFS~\cite{xu2021rethinking} & K400 & 25.9 \\
& TeG-FG & K400 & 27.7\\
& TeG-PS & K400 & \textbf{28.7}\\
\end{tabular}
\caption{\textbf{Spatiotemporal action localization on AVA-Kinetics.} TeG-PS outperforms its supervised pre-training counterpart by 8.9 mAP using the same R3D-50 backbone, as well as state-of-the-art unsupervised pre-training methods.}
\label{tab:avak-ft}
\end{table}

\section{Ablation Study}
We conduct ablation studies on a few key parameters in our proposed method. 
We use linear evaluation on VidSitu event classification to justify the performance on temporally fine-grained task and linear evaluation on Kinetics to represent video-level classification task. All experiments are conducted with 200 epochs of pre-training.

\begin{figure}[t]
 \centering
 \begin{subfigure}[b]{0.47\linewidth}
     \centering
     \includegraphics[width=\textwidth]{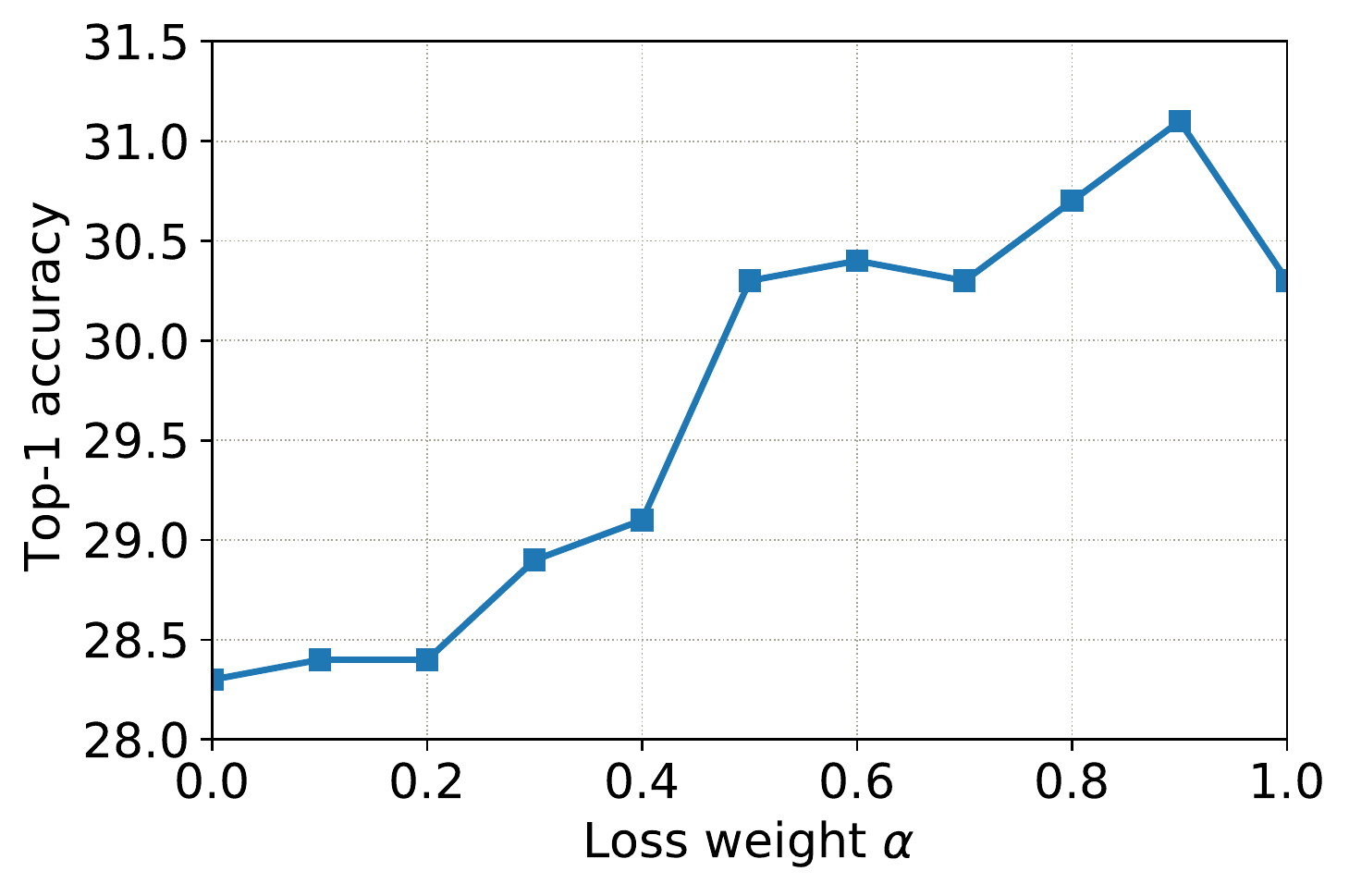}
     \caption{VidSitu event classification}
     \label{fig:ablation_alpha_vidsitu}
 \end{subfigure}
 \hfill
 \begin{subfigure}[b]{0.46\linewidth}
     \centering
     \includegraphics[width=\textwidth]{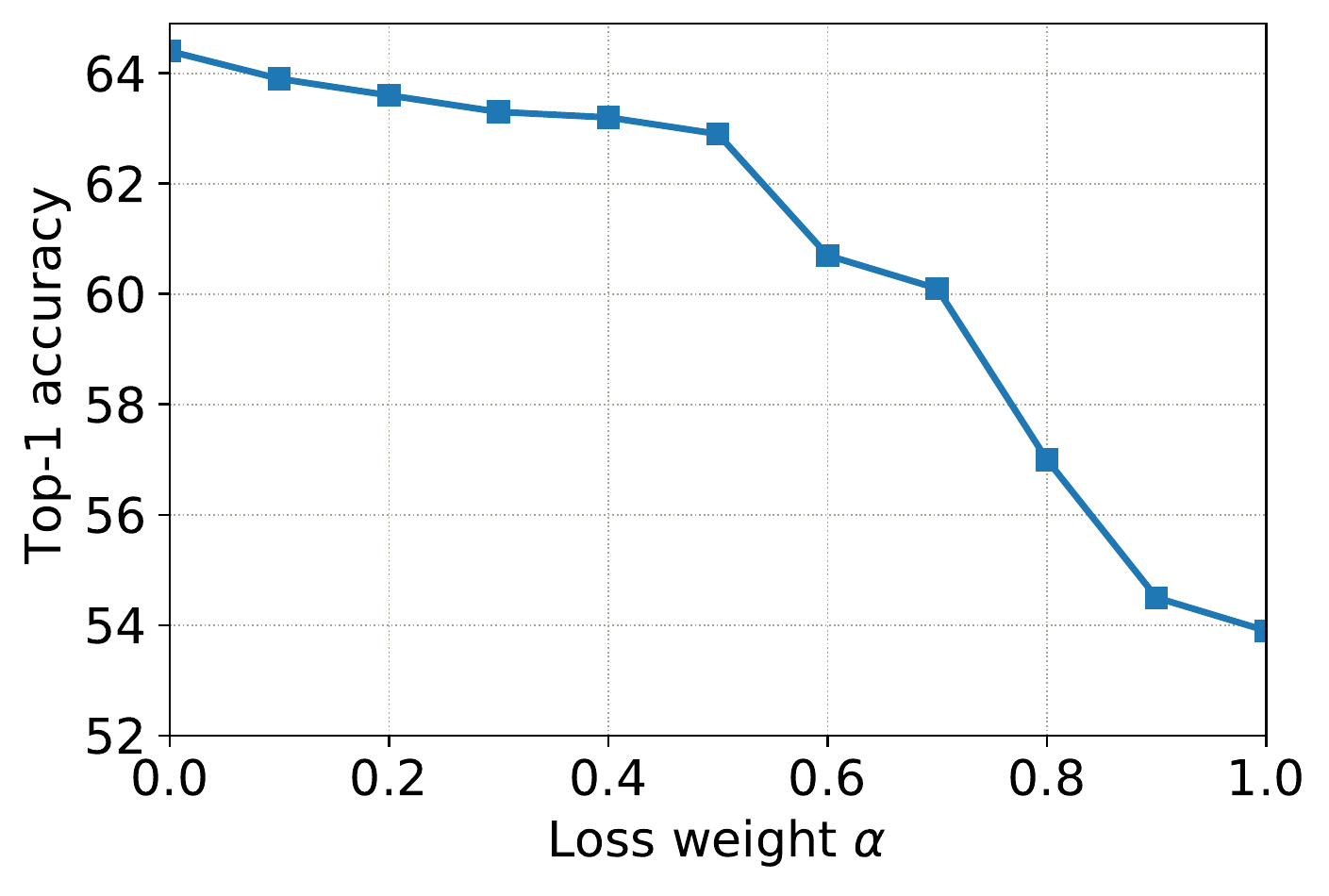}
     \caption{Kinetics action recognition}
     \label{fig:ablation_alpha_kinetics}
 \end{subfigure}
    \caption{\textbf{Ablation on loss weight $\alpha$.} VidSitu event classification and Kinetics action recognition require features of different granularities specified by $\alpha$.}
\vspace{-3mm}
    \label{fig:ablation_alpha}
\end{figure}

\paragraph{Loss weight.} 
Recall that in Equation~\ref{eq:final}, we propose to use a weight $\alpha$ to balance the learning of fine-grained and persistent loss. 
Intuitively, larger $\alpha$ would emphasize more on temporally fine-grained features and suppress the temporal persistency. 
We ablate the impact of $\alpha$ in Figure~\ref{fig:ablation_alpha}. 
On VidSitu (Figure~\ref{fig:ablation_alpha_vidsitu}), we observe that larger $\alpha$ generally yields better performance as expected except a performance drop from when $\alpha$ increases from 0.9 to 1.0.
This suggests that completely discarding the temporally persistent learning is not optimal. This is also the reason why we set $\alpha$ as 0.9 instead of 1.0 in TeG-FG. 
On Kinetics (Figure~\ref{fig:ablation_alpha_kinetics}), we see a consistent drop on the performance as $\alpha$ becomes larger. 
The reverse trend of performance further enhances our claim that different video tasks require features of different temporal granularities to achieve the best performance. 
Since we find bringing in temporally fine-grained features is harmful to Kinetics, we focus on VidSitu for the following ablation studies on parameters of temporally fine-grained learning.
To further demonstrate the effect of loss weight on learned features, We visualize the feature similarity of TeG-PS and TeG-FG in Appendix~\ref{app_visualization}.

\paragraph{Sampling strategy.} 
The proposed sampling strategy requires: 1) two clips to be asymmetric and 2) the short clip being contained in the long clip. 
We ablate on these two design choices in Table~\ref{tab:ablation_sample}. When two clips are both short, random sampling is identical to CVRL~\cite{qian2021spatiotemporal} and contained sampling losses the diversity in temporal context, thus resulting in poor performance. When two clips are asymmetric, random sampling still does not work well since the corresponding embeddings between the two clips are inaccurate in the cases that two clips do not have much overlap with each other. 

\begin{table}[h]
\small
\centering
\begin{tabular}{ccc}
clip $\backslash$ sampling & Random & Contained \\
\shline
Short - Short & 27.4 & 15.2 \\
Long - Short & 26.9 & \textbf{31.1}\\
\end{tabular}
\caption{\textbf{Ablation on sampling strategy.} The proposed samping of a long clip and a containing short clip performs the best. Results are on VidSitu event classification.} 
\label{tab:ablation_sample}
\end{table}

\begin{figure}[h]
  \centering
  \includegraphics[width=0.95\linewidth]{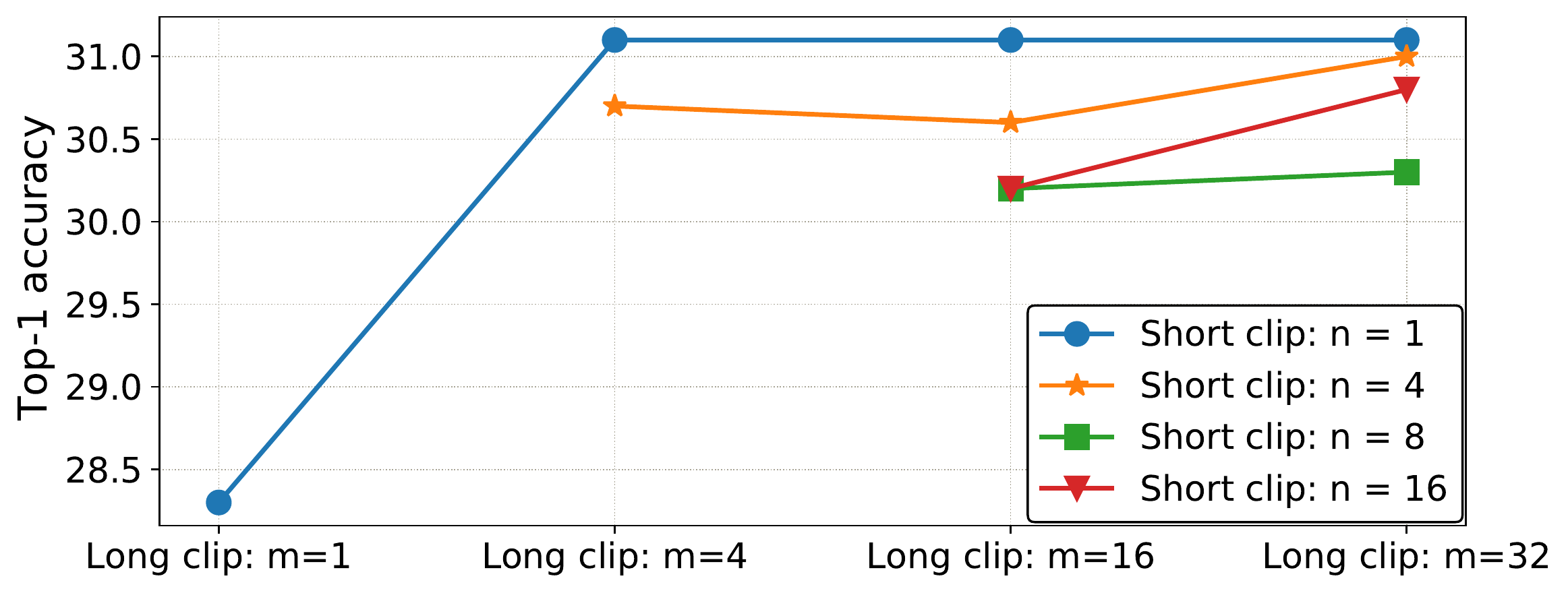}
  \caption{\textbf{Ablation on the choice of $n$ and $m$ in temporal aggregation}. Results are on VidSitu event classification.}
  \label{fig:ablation_nm}
\vspace{-2mm}
\end{figure}

\paragraph{Temporal aggregation.} 
The temporal aggregation parameters $m$ and $n$ determine how dense we want our fine-grained learning loss to be. 
We try different combinations of $m$ and $n$ and present their performances in Figure~\ref{fig:ablation_nm}. We choose $m=4, n=1$ as our default setting due to the simplicity and strong performance.

\section{Conclusions and Discussion}
\label{sec:conclusion}
This work studies the impact of temporal granularity in self-supervised video representation learning. 
We propose a flexible framework, which we call TeG, to learn video features of specified temporal granularity and observe that different video tasks require features of different temporal granularities.
This insight leads to state-of-the-art results on 8 video benchmarks.
We hope our study can inspire researchers in advancing video self-supervised learning. 

\paragraph{Limitations.} 
As observed from the experimental results, we find temporally fine-grained feature performs better on tasks like event classification and boundary detection, while temporally persistent feature shows great advantage on video-level action recognition and spatotemporal action localization. 
Manual effort is still needed to find the best recipe for different tasks. We hope our future work could extend TeG to learn a pyramid of representations with coarse to fine temporal granularities from unlabeled videos.
Then the learned representations can be easily transferred to downstream tasks in a more adaptive way.

\clearpage
{\small
\bibliographystyle{ieee_fullname}
\bibliography{egbib}
}

\clearpage
\begin{appendix}
\section{Additional Implementation Details}
\label{app_implementation}

\subsection{VidSitu}
\label{app_vidsitu}
VidSitu~\cite{Sadhu_2021_CVPR} contains 23.6k training, 1.3k validation and 1.3k test videos. Since the test set is held out for a challenge, we benchmark on the validation set. 
We download the videos with 720$\times$1280 resolution and 30 frame-per-second, using the script provided by the authors. 
During training, we apply random cropping with the area ratio set as (0.3, 1.0) and then resize frames to 224$\times$224 as in~\cite{qian2021spatiotemporal}.

\subsection{Kinetics-GEBD}
\label{app_gebd}
Kinetics-GEBD~\cite{shou2021generic} annotates 20k out of 240k training videos and all 20k validation videos from Kinetics-400. Each video is annotated by 5 sets of event boundaries. The multi-labeling does not affect our self-supervised pre-training stage since no labels are used. 

For generating training clips, we adopt the practice from the dataset authors by selecting the annotation entry with the highest F1 consistency score with other entries. 
The annotation is in the format of timestamps and we choose the closest frame to a ground truth timestamp as an event boundary. We adopt a 32-frame long sliding window with a stride of 3-frame. The 16$th$ frame of the sliding window is considered as the center frame, and the window would get a positive label when the time difference between the center frame and ground truth is less than 0.15 second. 

During fine-tuning, we sample a clip of 16 frames with a stride of 2 inside each window, and feed it into the video encoder. No temporal augmentation is applied. Instead of global average pooling, we conduct two separate average pooling before and after the center frame. We then concatenate the two features and perform binary classification. 

For prediction, we use the same sliding window and stride as in training. If a window is classified as positive, we use the timestamp of the center frame as the detected boundary. We would merge consecutive positive predictions into a single prediction by averaging their predicted timestamps.

Please refer to the original paper~\cite{shou2021generic} and the challenge evaluation code~\cite{gebd_eval} for details on how to deal with multiple ground truths and calculate the final F1 score.

\begin{figure}[t]
  \centering
  \includegraphics[width=0.95\linewidth]{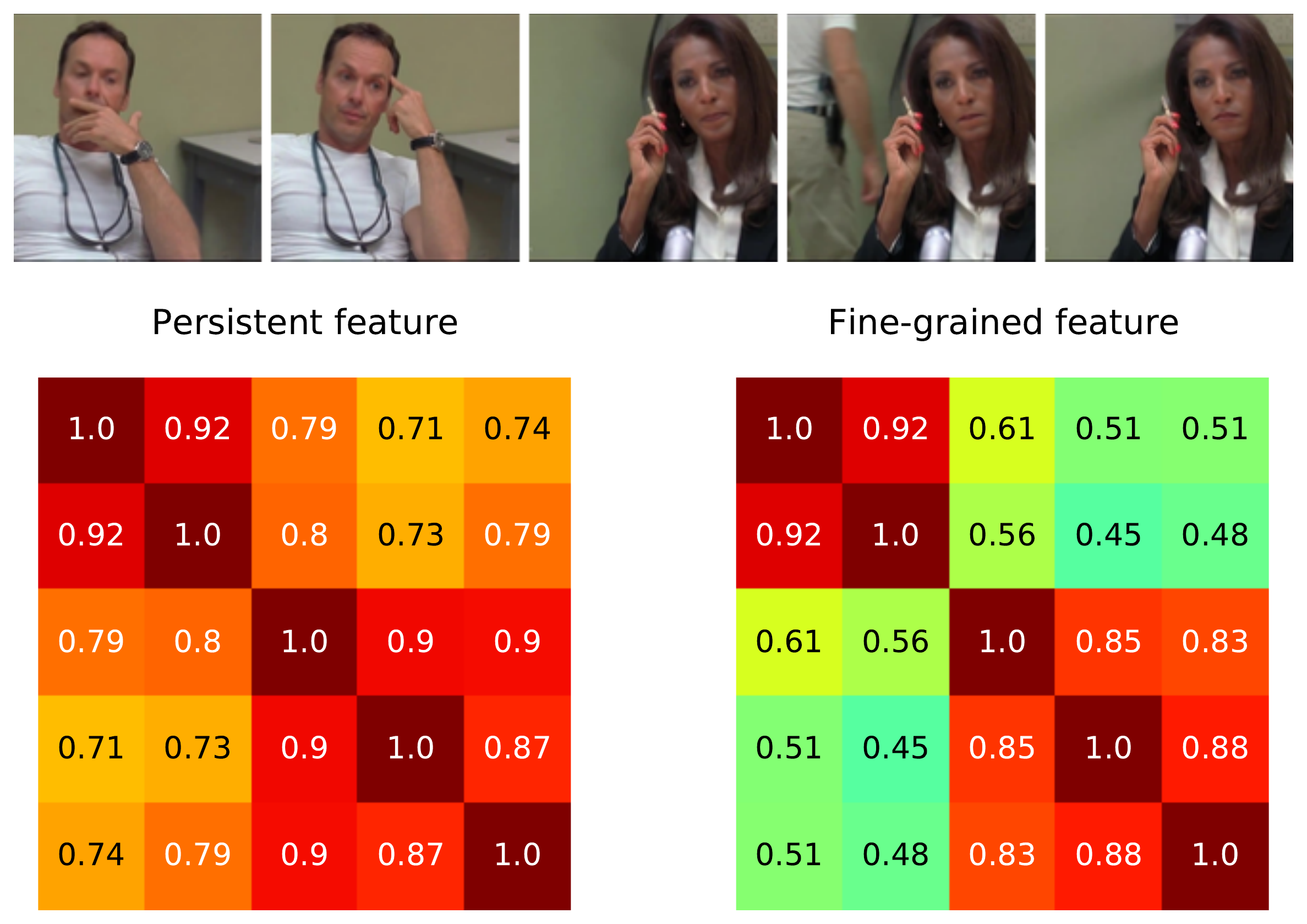}
  \caption{\textbf{Visualization of feature similarity}. Top row shows the center frame of each input clip. The left matrix is the similarity matrix of temporally persistent features and the right one comes from temporally fine-grained features.}
  \label{fig:visualization}
\end{figure}

\section{Visualization of Feature Similarity}
\label{app_visualization}

We provide a visualization of feature similarity to demonstrate the difference between temporally persistent and fine-grained features. 
Concretely, for every video in VidSitu validation set, we uniformly sample 5 consecutive clips and feed them into the trained video encoder to get their feature vectors. We then calculate the cosine similarities between all pairs of features, forming a 5$\times$5 similarity matrix. 
We extract learned features from two video encoders \textbf{TeG-PS} and \textbf{TeG-FG} introduced in Section~\ref{sec:experiments_implementation}.

Figure~\ref{fig:visualization} shows a representative example. 
Take the first row of the similarity matrix as an example, the 5 entries in the first row represent the feature similarity between (clip 1, clip 1), (clip 1, clip 2), $\cdots$, (clip 1, clip 5). 
Despite the great difference between two groups of clip 1-2 and clip 3-5, the temporally persistent features still obtain cosine similarities larger than 0.7. While for temporally fine-grained features, the similarities drop to around 0.45-0.6. 
This example clearly shows that temporally fine-grained features are more sensitive to the temporal content changes compared with temporally persistent features.

\begin{figure*}[t]
  \centering
  \setlength\tabcolsep{0.5pt}
  \renewcommand{\arraystretch}{5}
  \begin{tabular}{ccc}%
  \centering%
  \includegraphics[width=0.3\linewidth]{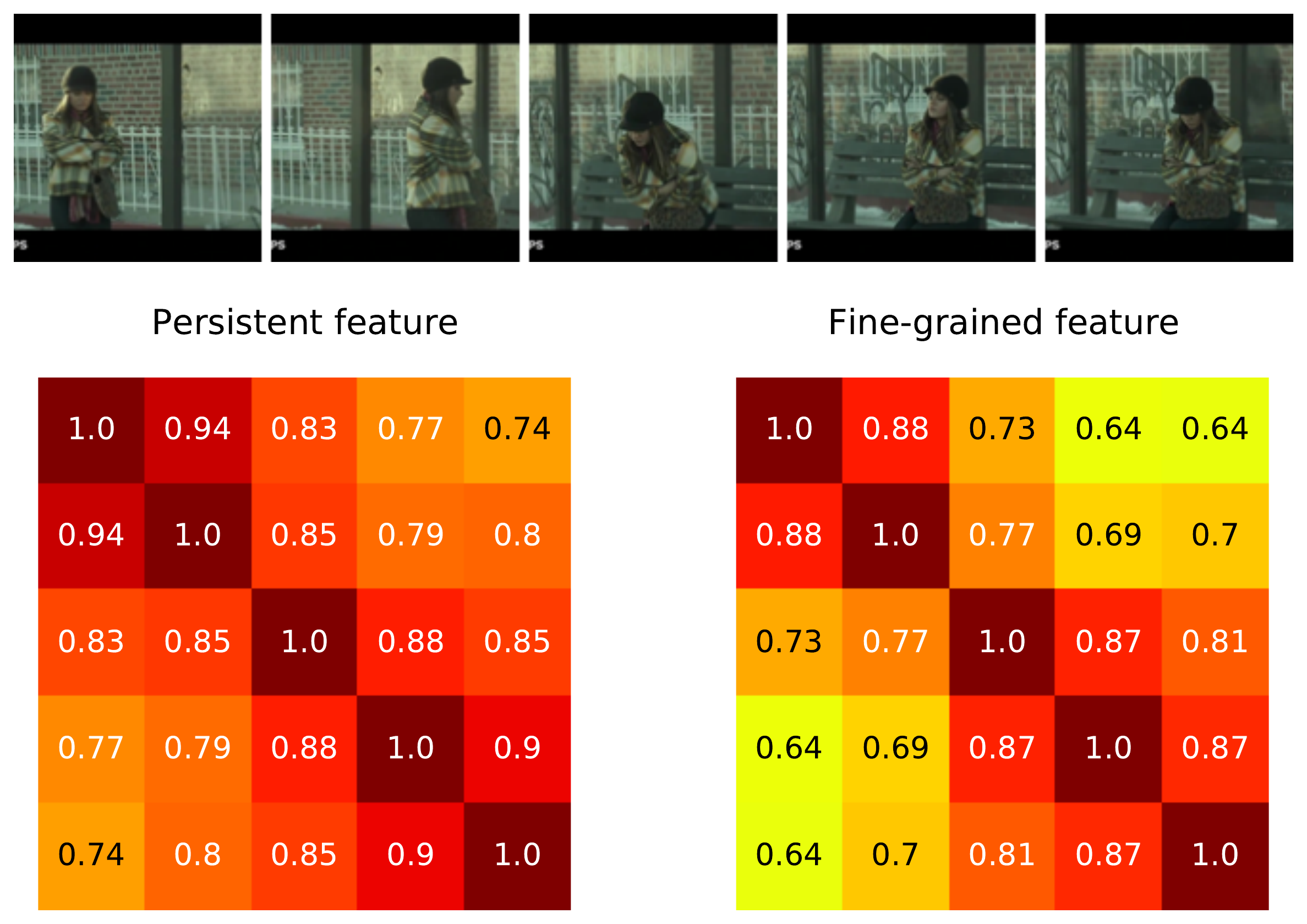}&~~~~~~
  \includegraphics[width=0.3\linewidth]{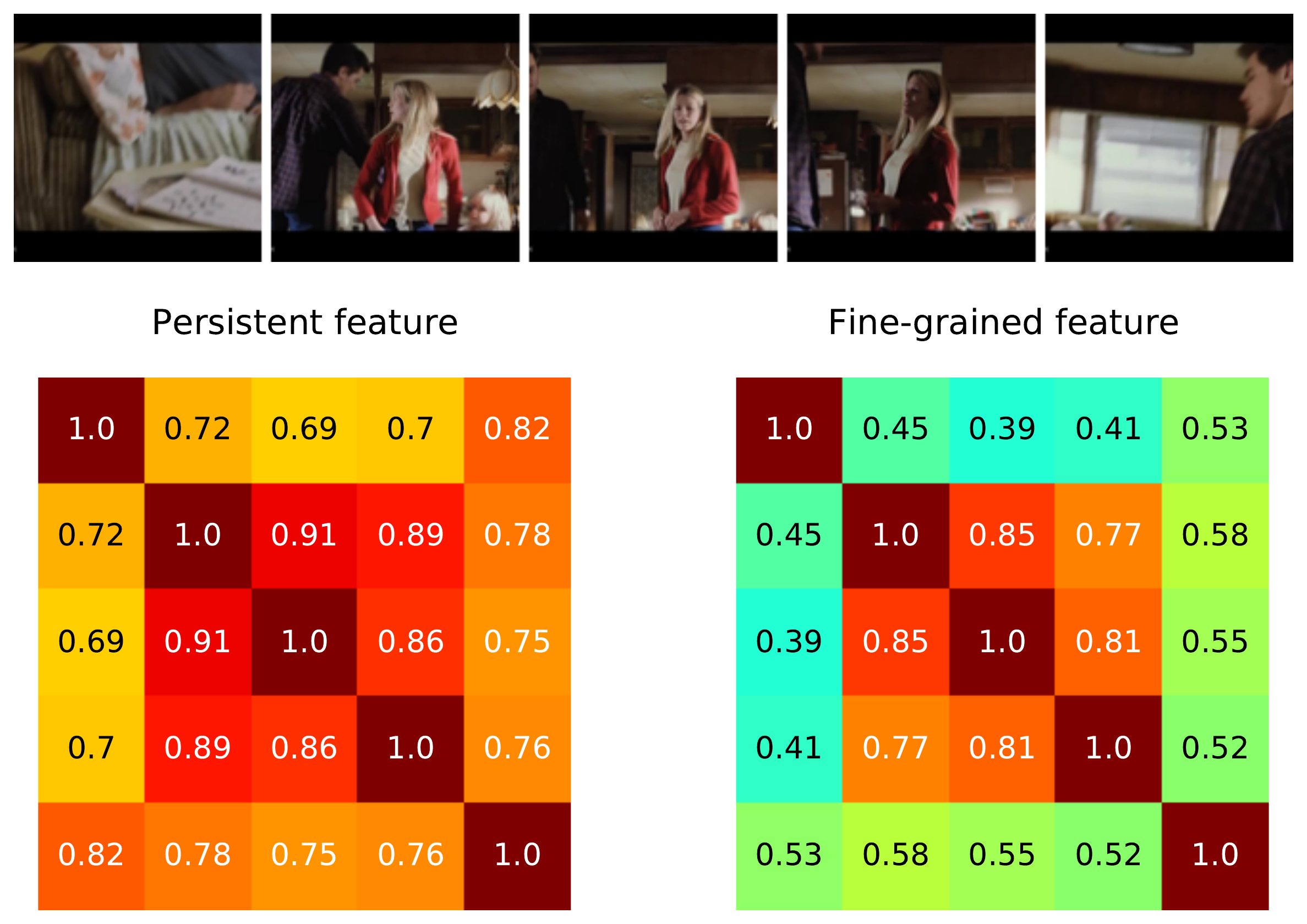}&~~~~~~
  \includegraphics[width=0.3\linewidth]{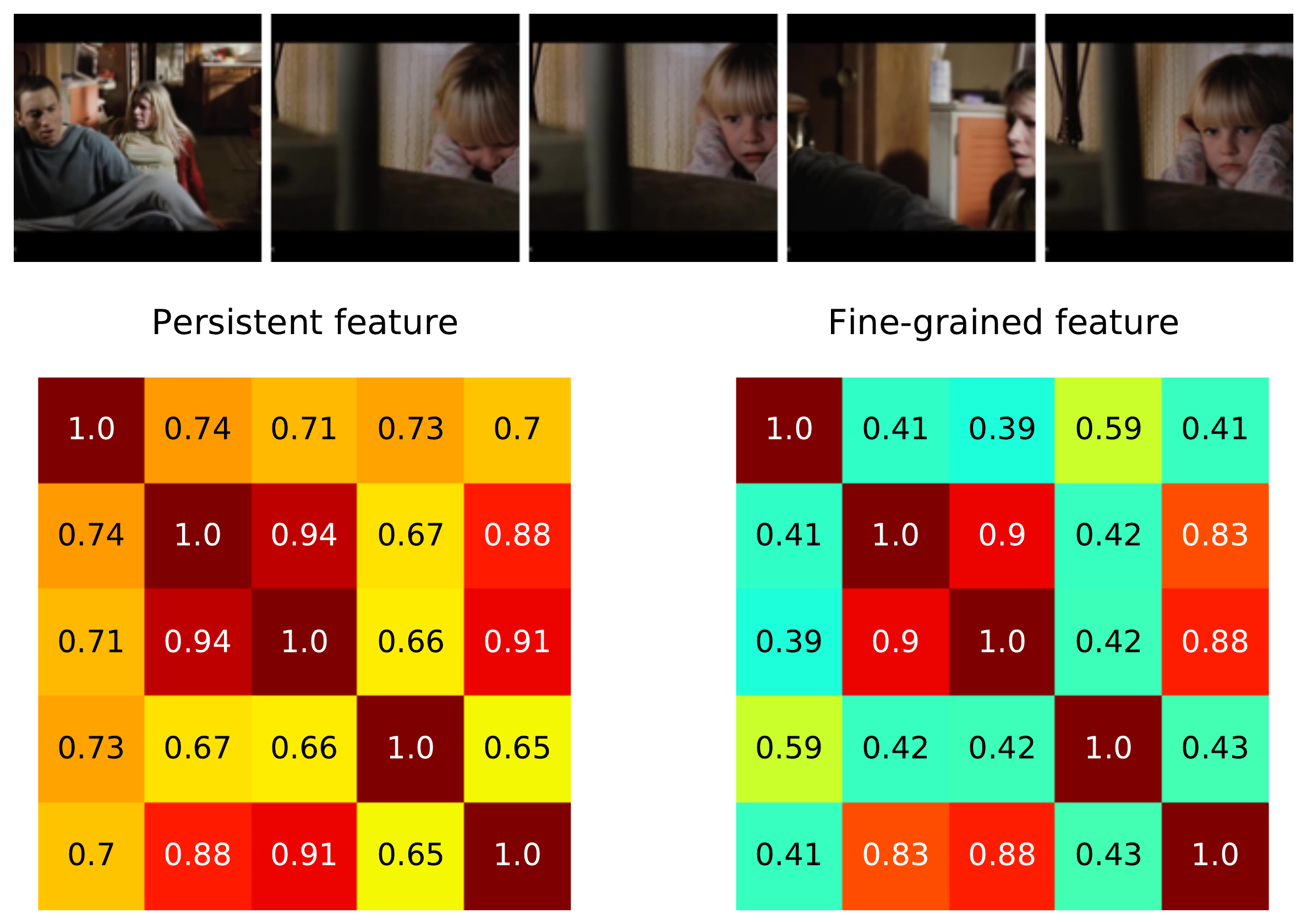}\\
  \includegraphics[width=0.3\linewidth]{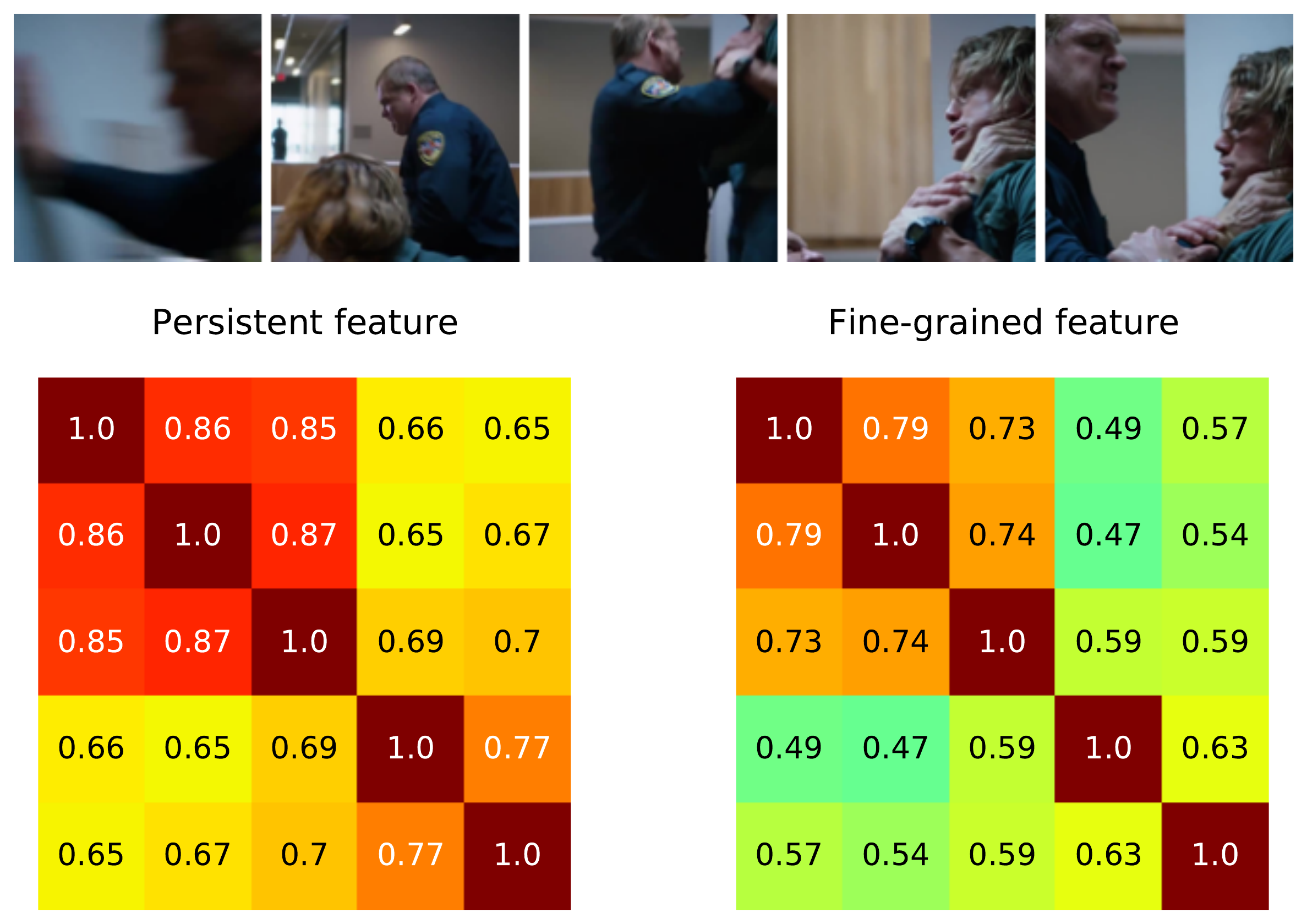}&~~~~~~
  \includegraphics[width=0.3\linewidth]{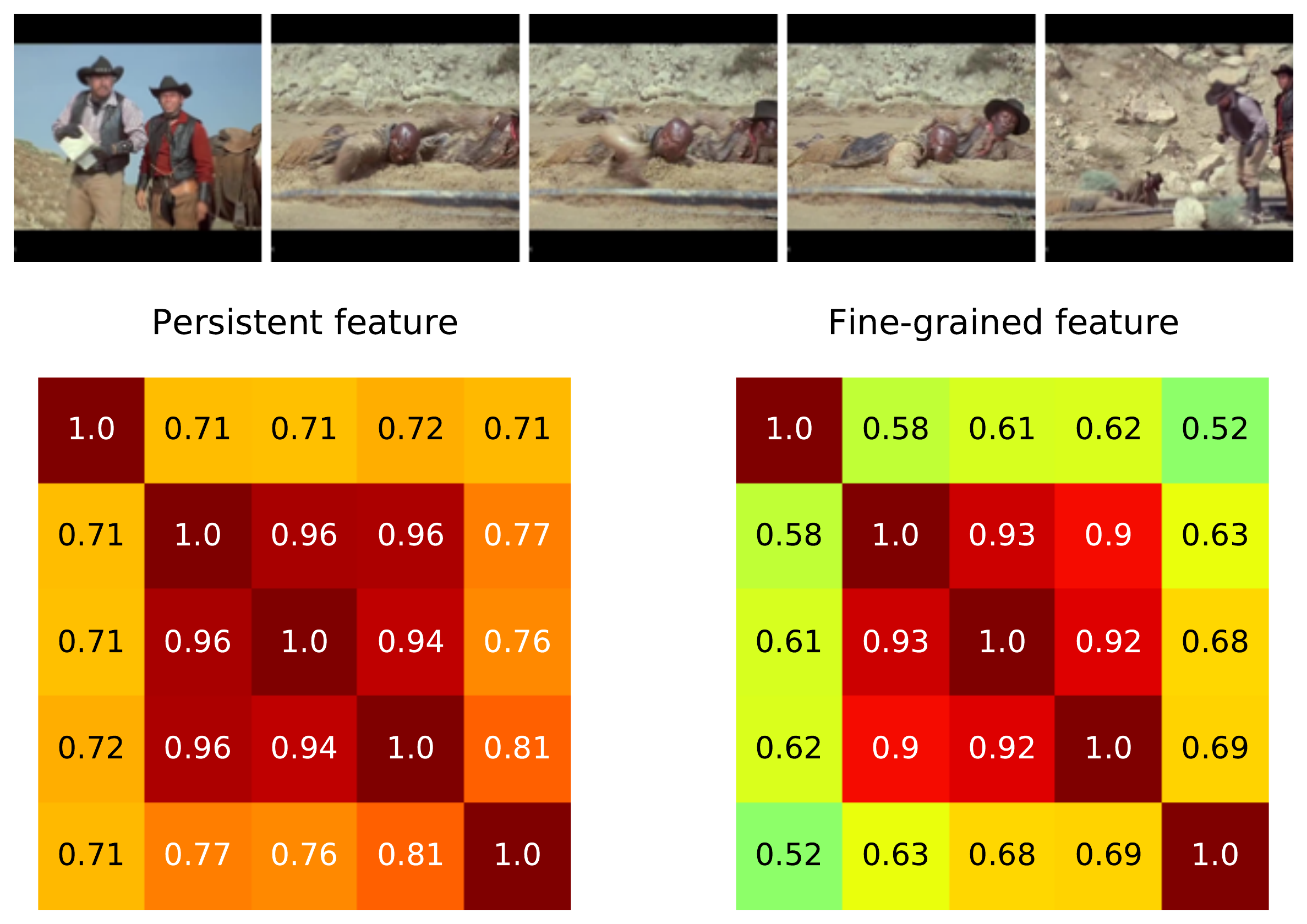}&~~~~~~
  \includegraphics[width=0.3\linewidth]{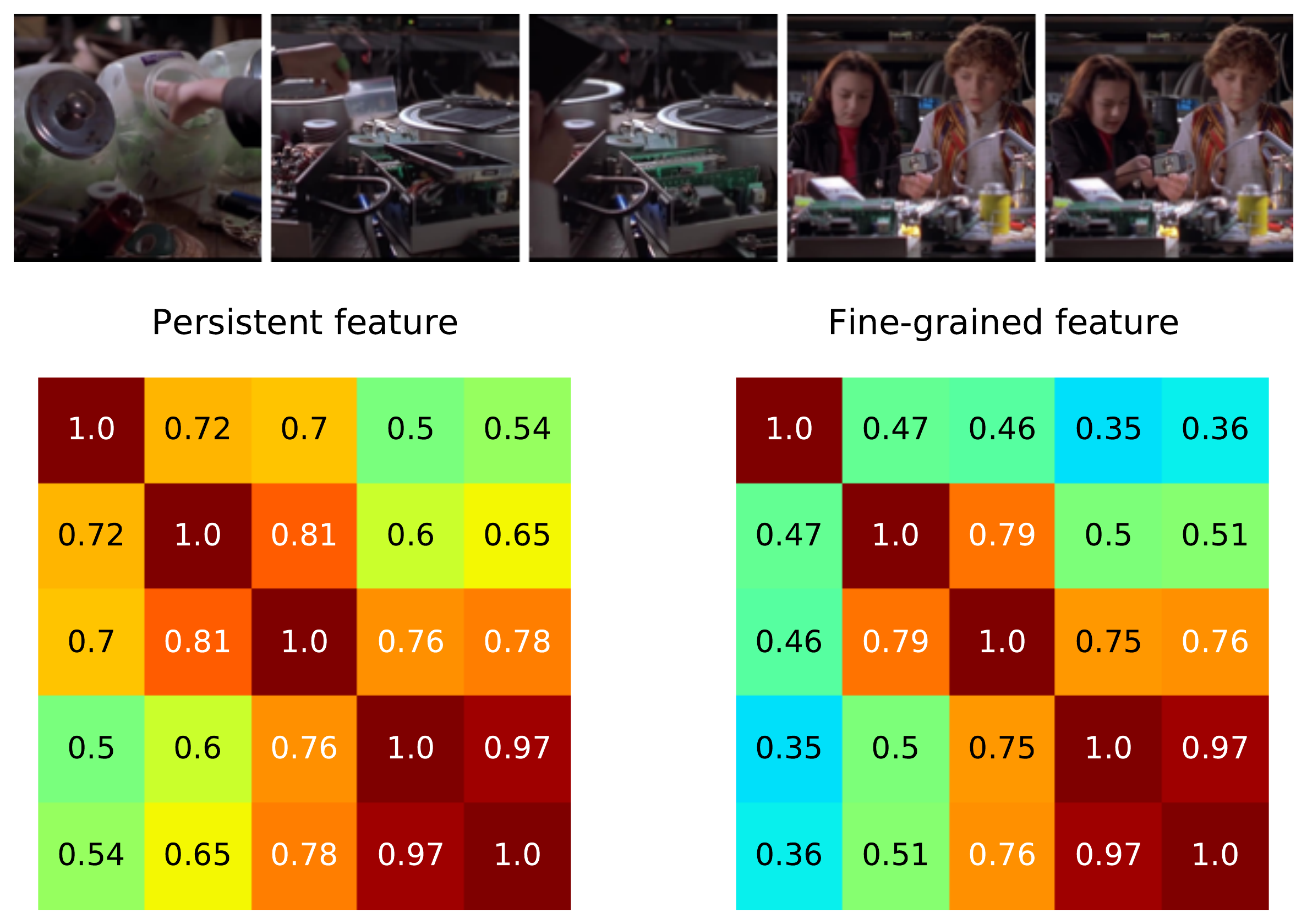}\\
  \includegraphics[width=0.3\linewidth]{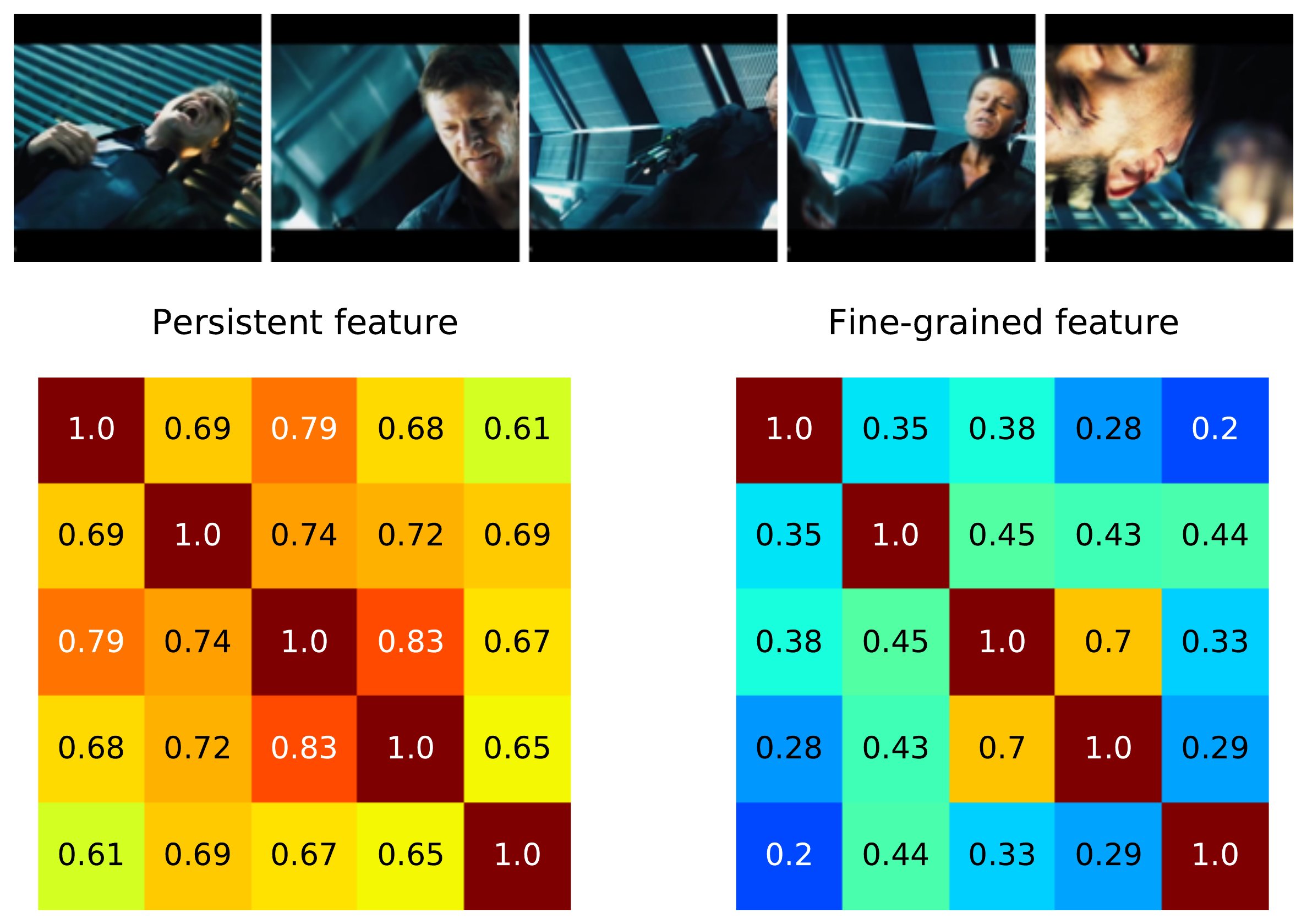}&~~~~~~
  \includegraphics[width=0.3\linewidth]{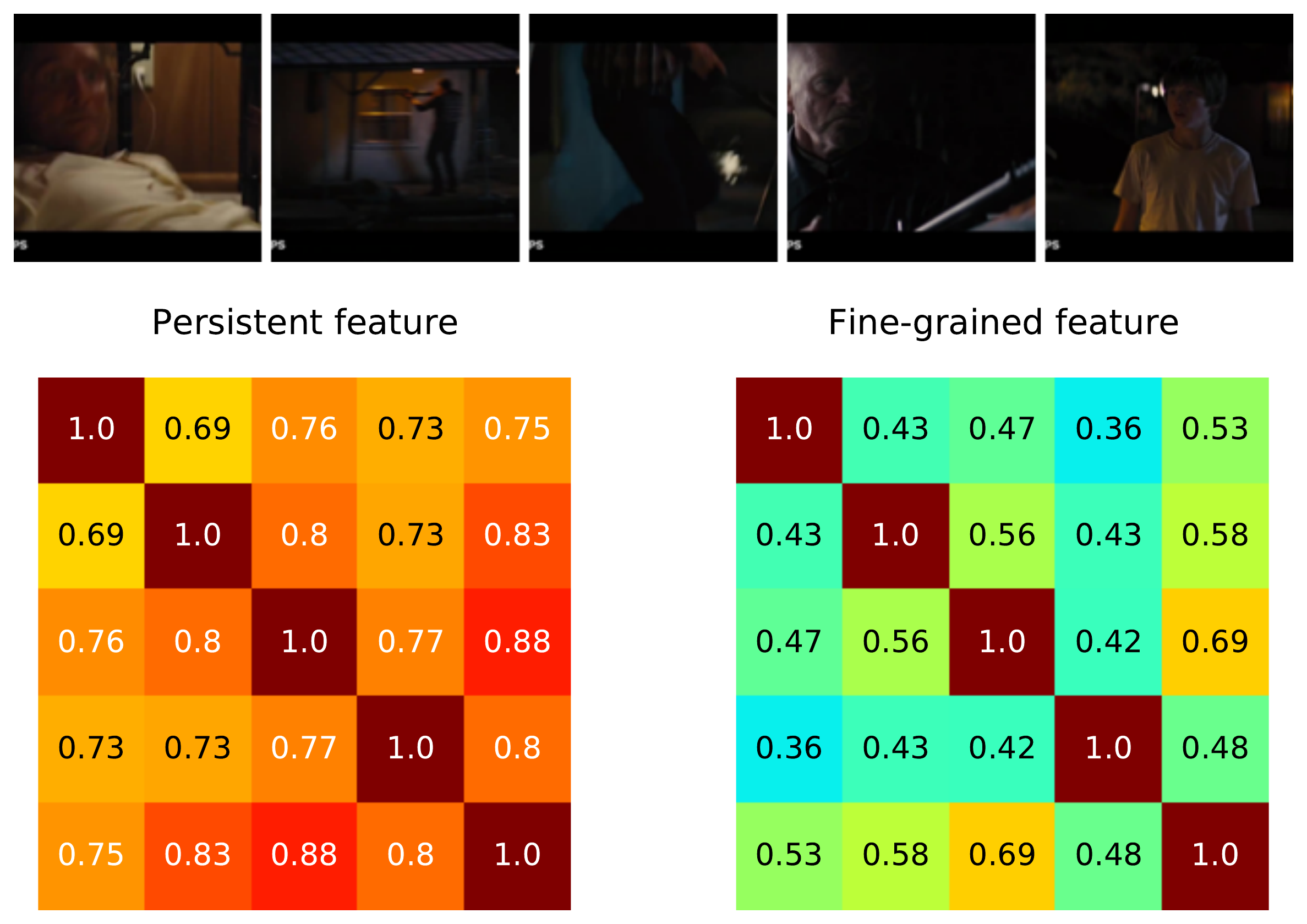}&~~~~~~
  \includegraphics[width=0.3\linewidth]{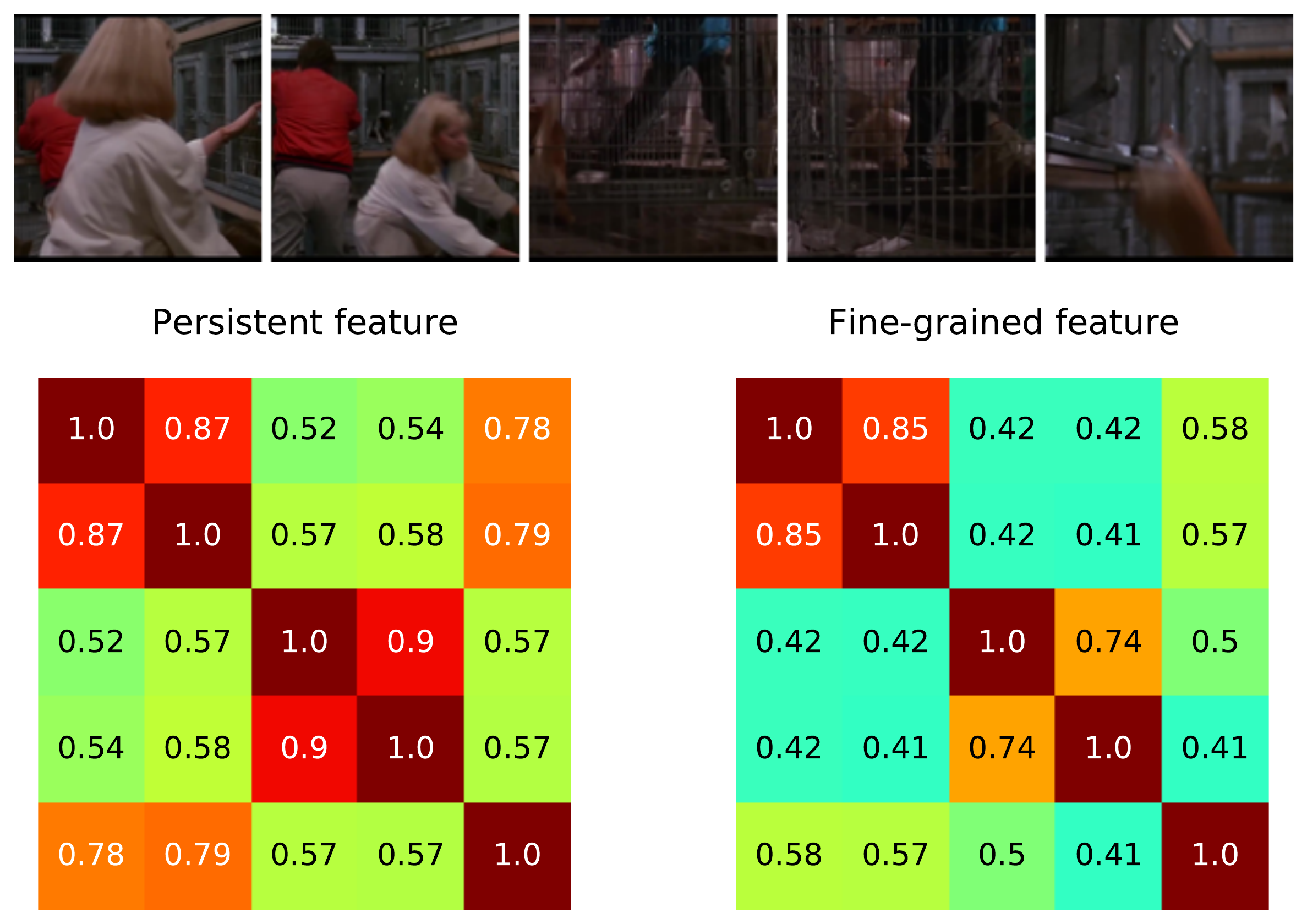}\\
  \includegraphics[width=0.3\linewidth]{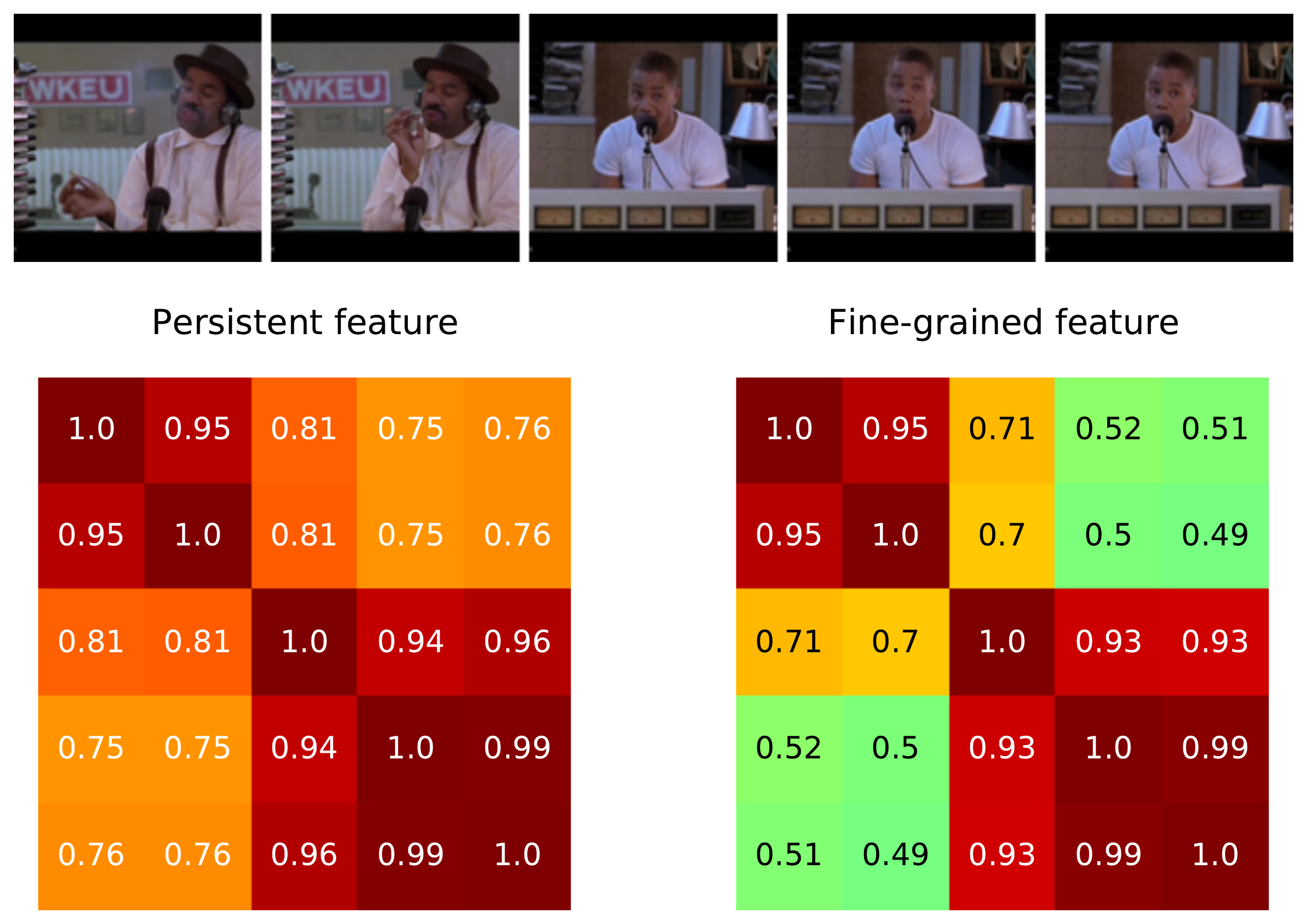}&~~~~~~
  \includegraphics[width=0.3\linewidth]{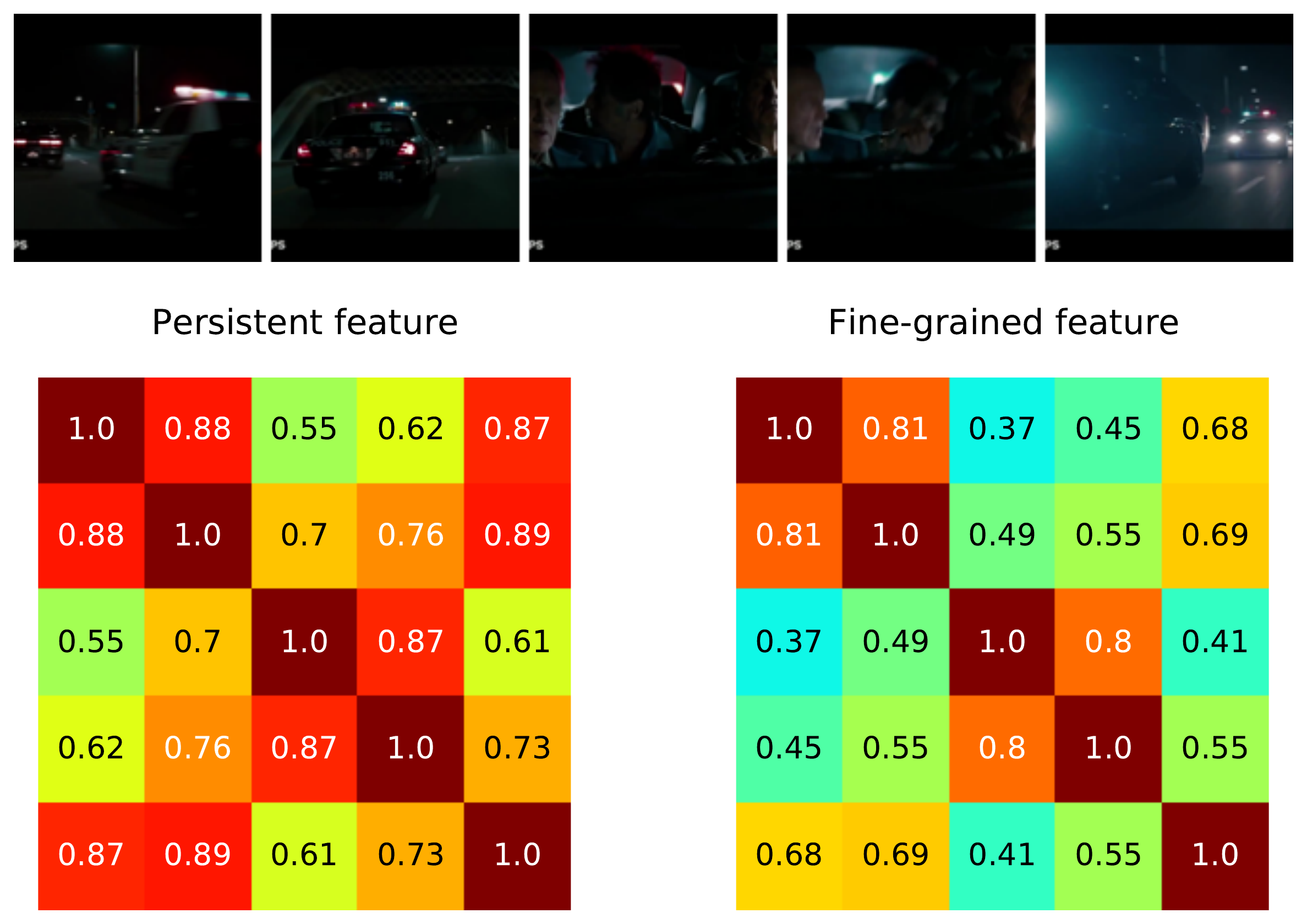}&~~~~~~
  \includegraphics[width=0.3\linewidth]{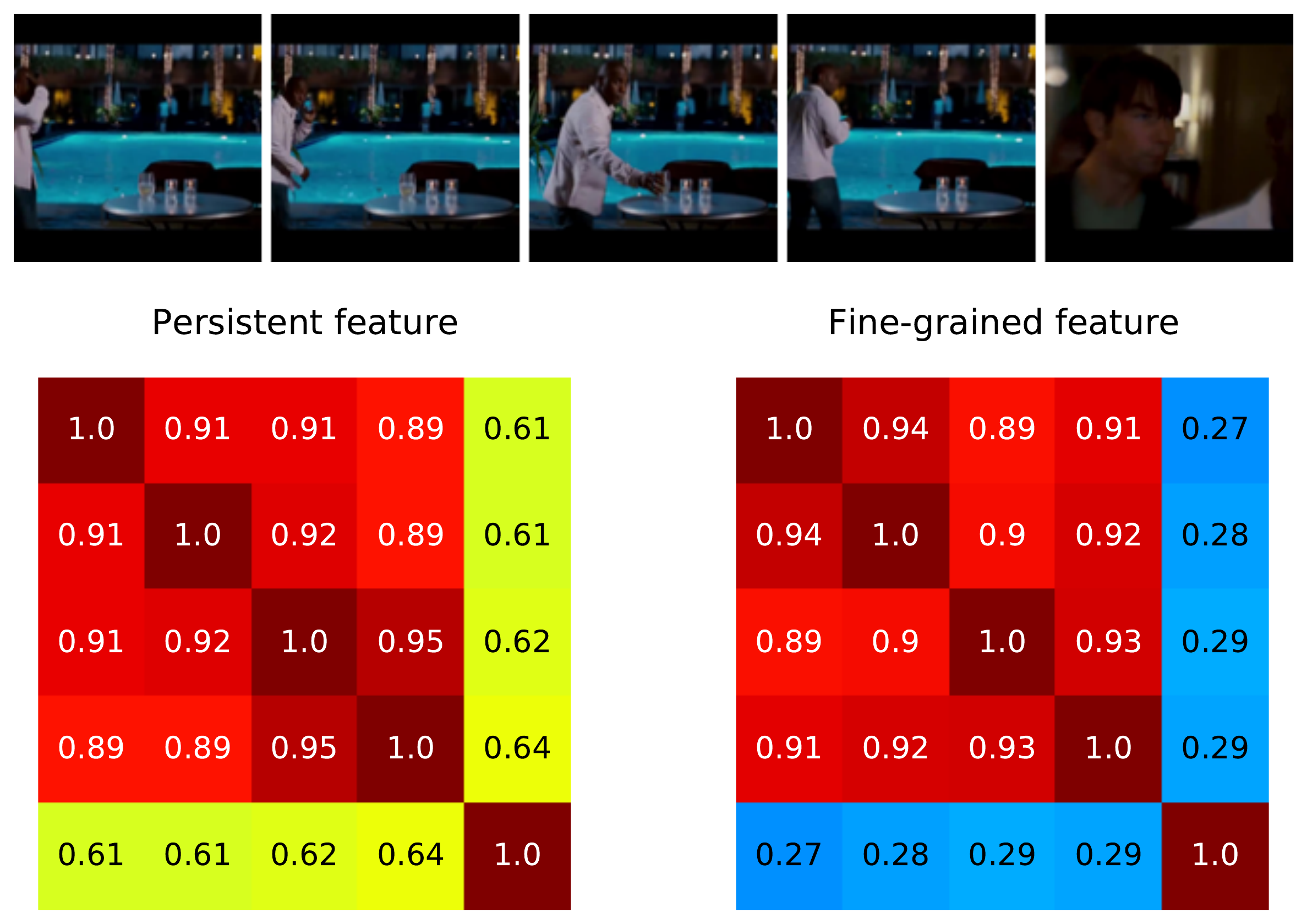}\\
  \includegraphics[width=0.3\linewidth]{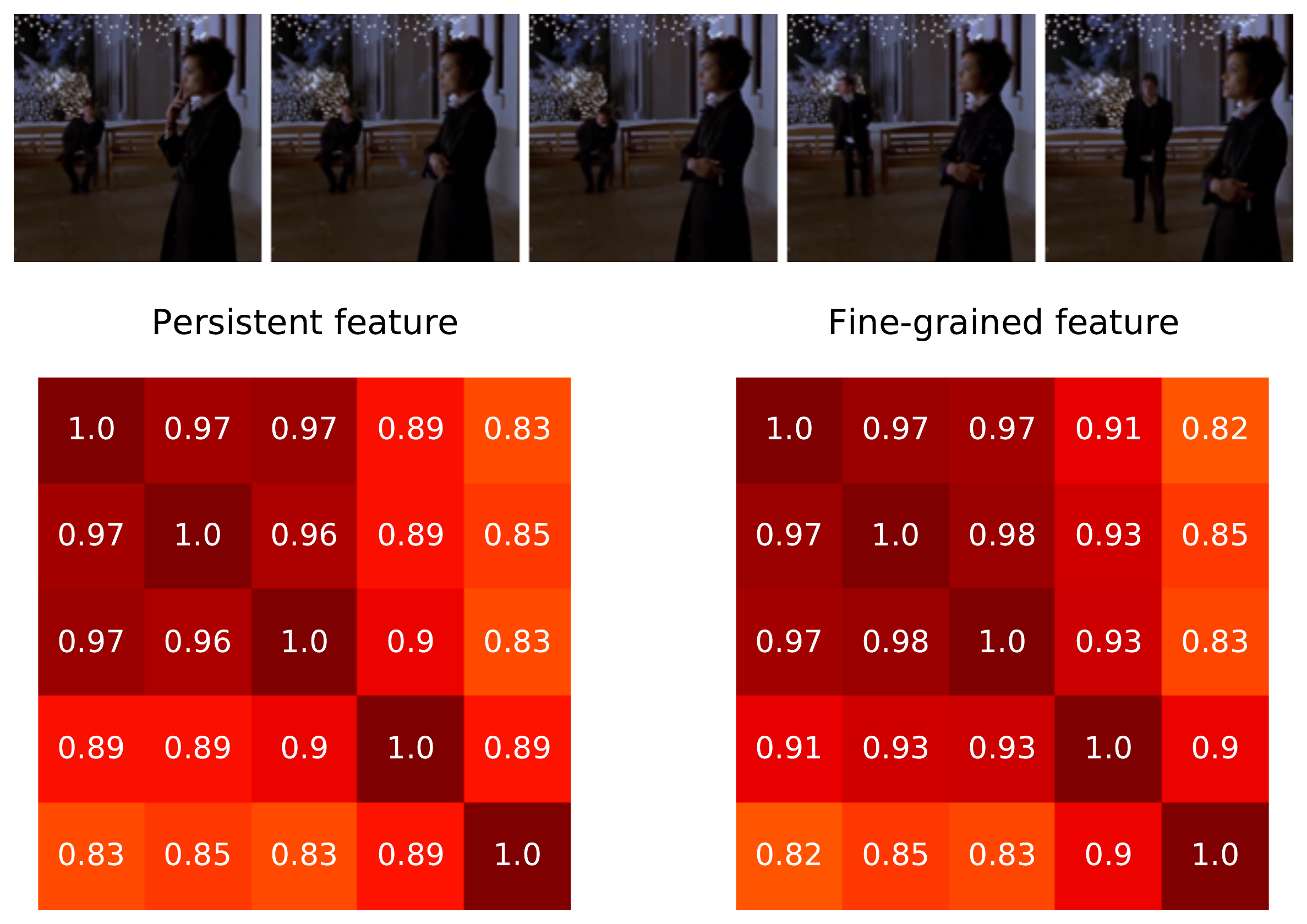}&~~~~~~
  \includegraphics[width=0.3\linewidth]{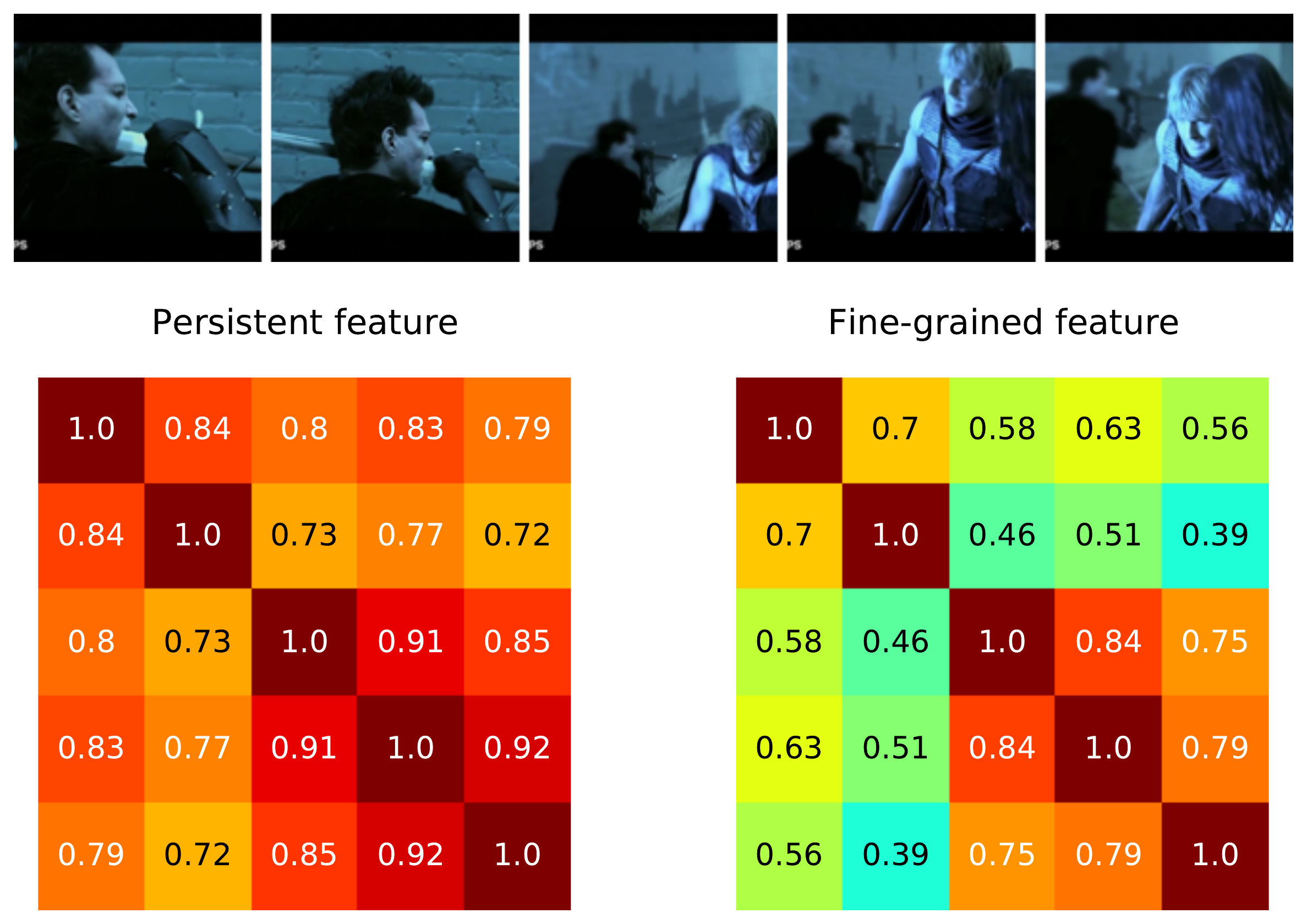}&~~~~~~
  \includegraphics[width=0.3\linewidth]{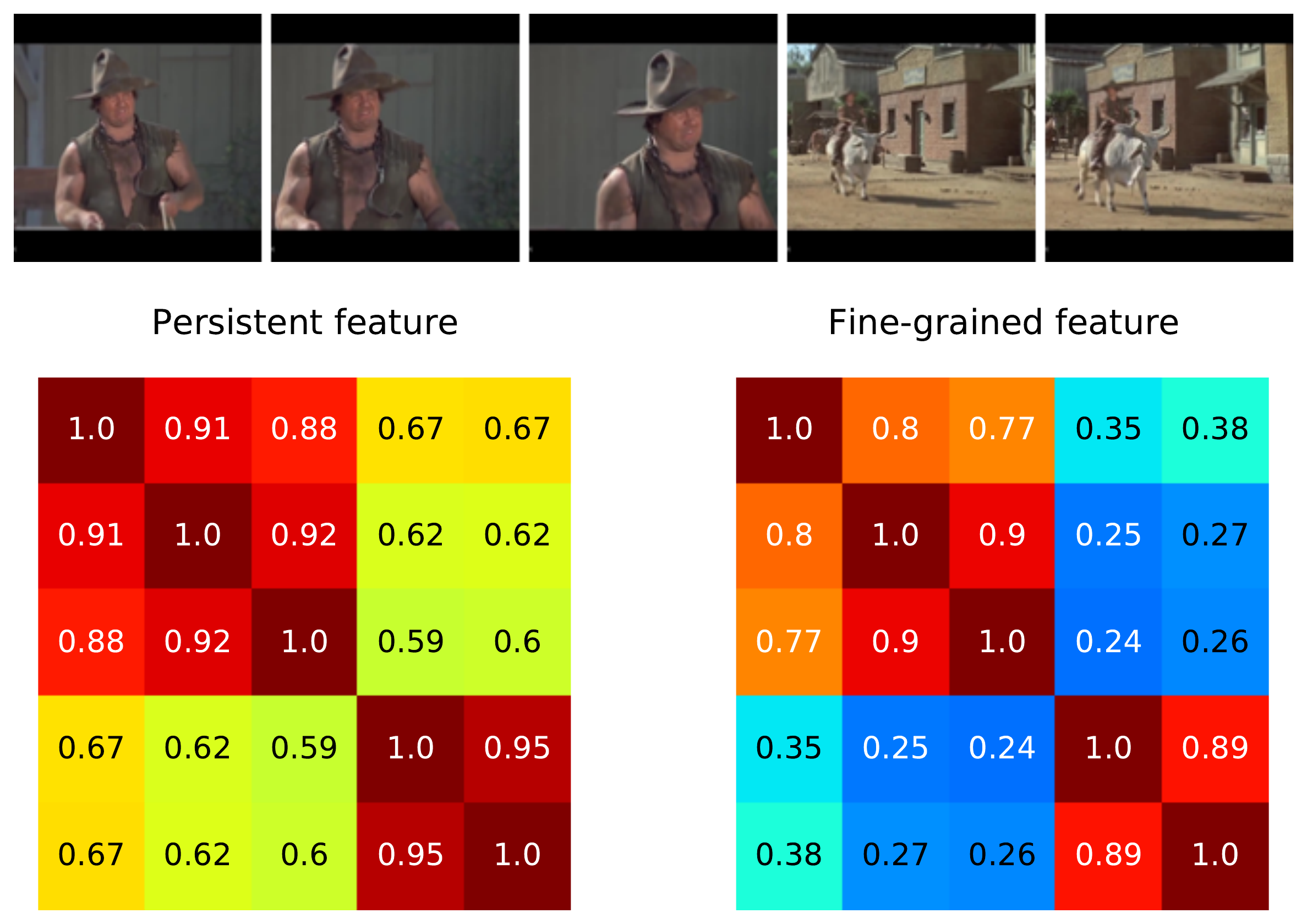}\\
\end{tabular}
\vspace{-4mm}
  \caption{\textbf{Random examples} of feature similarity on VidSitu validation videos. In each subfigure, we show the input video (top), the similarity matrices of temporally persistent features (bottom left) and temporally fine-grained features (bottom right).}
  \label{fig:more_visualization}
\end{figure*}

Additional randomly selected examples are shown in Figure~\ref{fig:more_visualization}.
From these examples, we can see that temporally persistent features would generally produce higher similarity scores compared with temporally fine-grained features.
We also draw a further observation that temporally fine-grained features are robust when the temporal content within the video changes very little (see the example in row 5, column 1 of Figure~\ref{fig:more_visualization}).

\end{appendix}
\end{document}